\documentclass{ecai}

\usepackage{booktabs}
\usepackage{algorithm}
\usepackage{algpseudocodex}
\usepackage{amsmath,amssymb,amsfonts,amsthm}
\usepackage[inline]{enumitem}
\usepackage{multirow}
\usepackage{bm}
\usepackage[subpreambles=true]{standalone}
\usepackage[caption=false]{subfig}
\usepackage{nicefrac}
\usepackage{mystyle}
\usepackage{xr}
\usepackage{esvect}

\usepackage{soul}

\newtheorem{definition}{Definition}
\newtheorem{observation}{Observation}

\newcommand{\BibTeX}{B\kern-.05em{\sc i\kern-.025em b}\kern-.08em\TeX}

\newcommand{\method}{Continuous Imitation Learning from Observation}
\newcommand{\abrev}{CILO}

\newcommand{\ie}{{\it i.e.}}
\newcommand{\eg}{{\it e.g.}}

\newcommand{\tuple}[1]{\ensuremath{\langle{#1}\rangle}}

\begin{document}


\begin{frontmatter}

\paperid{278}


\title{Explorative Imitation Learning: A Path Signature Approach for Continuous Environments}


\author[A]{
    \fnms{Nathan}~\snm{Gavenski}
    \thanks{Corresponding Author. Email: nathan.schneider\_gavenski@kcl.ac.uk}
}

\author[]{\fnms{Juarez}~\snm{Monteiro}}
\author[B,C]{\fnms{Felipe}~\snm{Meneguzzi}}
\author[D]{\fnms{Michael}~\snm{Luck}}
\author[A]{\fnms{Odinaldo}~\snm{Rodrigues}} 

\address[A]{King's College London, London, United Kingdom}
\address[B]{University of Aberdeen, Aberdeen, United Kingdom}
\address[C]{Pontif{\'i}cia Universidade Cat{\'o}lica do RS, Porto Alegre, Brazil}
\address[D]{University of Sussex, Sussex, United Kingdom}


\begin{abstract}
Some imitation learning methods combine behavioural cloning with self-supervision to infer actions from state pairs.
However, most rely on a large number of expert trajectories to increase generalisation and human intervention to capture key aspects of the problem, such as domain constraints.
In this paper, we propose \method{}~(\abrev), a new method augmenting imitation learning with two important features:
(i) exploration, allowing for more diverse state transitions, requiring less expert trajectories and resulting in fewer training iterations; and
(ii) path signatures, allowing for automatic encoding of constraints, through the creation of non-parametric representations of agents and expert trajectories.
We compared \abrev{} with a baseline and two leading imitation learning methods in five environments.
It had the best overall performance of all methods in all environments, outperforming the expert in two of them.
\end{abstract}

\end{frontmatter}


\section{Introduction} 
\label{sec:introduction}

One of the most common forms of learning is by watching someone else perform a task and, afterwards, trying it ourselves.
As humans, we can observe an action being performed and transfer the acquired knowledge into our reality.
In this respect, it is less challenging to achieve a goal in an optimal way by observing how an expert behaves; in the field of computer science, this is Imitation Learning (IL).
Unlike conventional reinforcement learning, which depends on a reward function, IL learns from expert guidance, and is concerned with an agent's acquisition of skills or behaviours by observing a `teacher' perform a given task. 

\newcommand{\otr}[1]{{\color{orange}#1}}
\textit{Learning from demonstration} is the obvious approach for IL, requiring expert demonstrations, which are `trajectories' including actions performed along the way to goal completion~\citep{HusseinEtAl2017}.
Such an approach uses the trajectories to learn an approximate policy that behaves like the expert.
Learning from demonstration suffers from two significant drawbacks in practice: poor generalisation in environments with multiple alternative trajectories that achieve a goal, which is bound to occur when the dataset size increases, and the unavailability of data about the expert's actions.
\textit{Learning from observation} (LfO) overcomes these limitations by learning a task without direct action information via self-supervision, which increases generalisation~\cite{Gavenski2022how}. 
This allows a model to learn from sample executions without action information, which would otherwise be unusable. 
LfO approaches often rely on techniques from classification to improve sample-efficiency~\citep{zhu2020off} and generalisation~\citep{monteiro2020augmented}.
Such agents require fewer expert trajectories, yielding more general approaches that are, hence, adaptable to unseen scenarios.
However, these methods still fail to leverage some useful learning features, particularly the use of an exploration mechanism.

Some existing work~\citep{EdwardsEtAl2019,gavenski2020imitating} requires manual intervention in different stages of the process, e.g., the hard-coding of environment goals, which is not feasible in complex environments, such as robotic systems with multifaceted goals.
Other work~\citep{gavenski2020imitating,kidambi2021mobile,zhu2020off} is limited in that learning the environment dynamics depends strongly on previously collected samples that usually do not relate to how the environment dynamics operate under expert behaviour, such as random transitions, or prior knowledge of the dynamics of environments.
In addition, maintaining self-supervision~\citep{gavenski2020imitating,kidambi2021mobile} for an IL method is important since unlabelled data is more readily available, \eg, from sources that are not necessarily meant for agent learning.

In this paper, we propose a novel LfO approach to IL called \method~(\abrev)  that addresses the above issues. 
\abrev{} 
\begin{enumerate*}[label=(\roman*)]
    \item eliminates the need for manual intervention when using different environments by discriminating between policy and expert;
    \item requires fewer samples for learning by leveraging exploration and exploitation; and 
    \item does not require expert-labelled data, thus remaining self-supervised.
\end{enumerate*}
We evaluated \abrev{} in five widely used continuous environments against a baseline and two leading LfO methods (see Section~\ref{sec:experimental}). 
Our results show that \abrev{} outperformed all of the alternatives, surpassing the expert in two of five environments.

\abrev{}'s new mechanisms are model-agnostic and applicable to a wider range of environment dynamics than those of the compared LfO alternatives.
We argue that the new mechanisms can be readily incorporated into other IL methods, paving the way for more robust and flexible learning techniques.

\section{Problem Formulation} \label{sec:sub:problem}

We assume the environment to be an MDP $M = \tuple{S, A, T, r, \gamma}$, in which $S$ is the state space, $A$ is the action space, $T$ is a transition model, $r$ is the {immediate} reward function, and $\gamma$ is the discount factor \citep{SuttonBarto1998}. 
Although in general an MDP may carry information regarding the reward and discount factors, we consider that this information is inaccessible to the agent during training, and the learning process does not depend on it.
Solving an MDP yields a policy $\pi$ with a probability distribution over actions, giving the probability of taking an action $a$ in state $s$. We denote the expert policy by $\teacher$.

A common self-supervised approach to solving a task via IL uses an Inverse Dynamic Model $\mathcal{M}$. 
$\mathcal{M}$ uses a set of state transition samples $(s_t,s_{t+1})$ to predict the action performed in the transitions.
By training $\mathcal{M}$ to infer the actions in the state transitions, these approaches can automatically annotate all expert trajectories $\Tau^\teacher$ with actions without the need for human intervention~\citep{TorabiEtAl2018,monteiro2020augmented,gavenski2020imitating}.
The agent policy $\agent$ then uses these \textit{self-supervised} expert-labelled states ($s^\teacher$, $\hat{a}$) to learn to predict the most likely action given a state $P(a \mid s^\teacher)$. 
\citeauthor{TorabiEtAl2018}~\cite{TorabiEtAl2018} show that applying an iterative process in self-supervised IL approaches helps $\agent$ achieve better performance.
Initially, $\mathcal{M}$ uses only single transition samples $I^{pre}$ from $\agent$ and its randomly initialised weights.
At each iteration, these approaches use $\agent$ to create new samples $I^{pos}$ that are used to fine-tune $\mathcal{M}$.
However, using all transitions from $\agent$ makes this iterative approach susceptible to getting stuck in \textit{local minima} due to class imbalance from the $I^{pos}$ data.
\citeauthor{monteiro2020augmented}~\cite{monteiro2020augmented} propose a solution that introduces a goal-aware function to sample from all trajectories at each epoch.
This function does not require an aligned goal from the environment, hence it is up to the user to choose a desired goal.
If $\agent$ reaches this goal, the trajectory will be used.
Finally, it is sensible to assume that $\mathcal{M}$ is not well-tuned during early iterations and predicts mostly wrong labels.
Therefore, \citeauthor{gavenski2020imitating}~\cite{gavenski2020imitating} implement an exploration mechanism that uses the softmax distribution of the output as weights to sample actions proportionally to optimality from the model's prediction. 
As the confidence in the model increases, it predicts suboptimal actions less than the \textit{maximum a posteriori estimation}.
By exploring using the model's confidence, their approach can learn under the exploration and exploitation phases, helping $\mathcal{M}$ to converge faster.
Nevertheless, creating a handcrafted goal-aware function and using a softmax distribution as an exploration mechanism requires manual intervention and discrete actions. As a result, these methods become unsuitable for more complex environments where goal achievement is non-trivial to check.

\section{\method} \label{sec:method}

\begin{figure}[b]
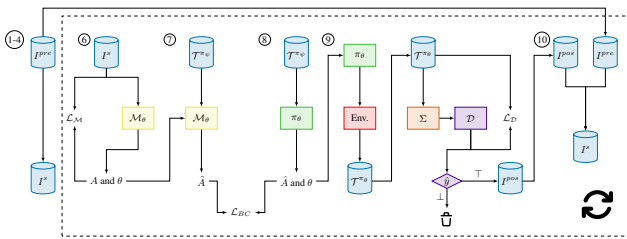

    \centering
    \includestandalone[width=.95\columnwidth]{content/figures/diagram}
    \caption{\abrev's training cycle.}
    \label{fig:pipeline}
\end{figure}

We address the need for manual intervention and for maintaining self-supervision in \abrev{} through two key innovations: an exploration mechanism used when the action predictions are uncertain; and a discriminator to interleave random and current states to improve the prediction of self-supervised actions.
\abrev{} achieves this by employing three different models: 
\begin{enumerate*}[label=(\roman*)]
    \item the inverse dynamic model $\mathcal{M}$ to predict the action responsible for a transition between two states $P(a \mid s_t, s_{t+1})$;
    \item a policy model $\agent$ that uses the self-supervised labels $\hat{a}$ to imitate the expert $\teacher$ given a state $P(a \mid s_t)$; and
    \item a discriminator model $\mathcal{D}$ to discriminate between $\teacher$ and $\agent$, creating newer samples for $\mathcal{M}$.
\end{enumerate*}

Algorithm~\ref{algo:method} provides an overview of \abrev{}'s learning process.
First, \abrev{} initialises all models with random weights and uses the random initialised policy to collect random samples $I^{pre}$ from the environment (Lines~\ref{alg:line:initialize}-\ref{alg:line:ipre}).
The dynamics model uses these random samples to train in a supervised manner (Function $\Call{trainM}{}$, Line~\ref{alg:line:idm}).
These samples are vital since they help $\mathcal{M}$ learn how actions cause environmental transitions without expert behaviour-specific knowledge.
$\Call{trainM}{}$ uses the loss from Eq.~\ref{eq:lm}, where $\theta$ are the model's current parameters, $S$ is the vector for state representations, $A$ is the action vector representation, and $t$ is the timestep.
\begin{equation} \label{eq:lm}
    \mathcal{L}_{\mathcal{M}}(I^s, \theta) 
        = \sum_{t=1}^{I^s}
        \left | \mathcal{M}_\theta(S_t, S_{t+1}) - A_t \right |
\end{equation}
\noindent
With the updated parameters $\theta$, $\mathcal{M}$ predicts the self-supervised labels $\hat{A}$ to all expert transitions ($\mathcal{T}^{\teacher}$ in Line~\ref{alg:line:self}).
\abrev{} then uses these expert labelled transitions to train $\agent$ using behaviour cloning (Function $\Call{BehaviouralCloning}{}$ with Eq.~\ref{eq:lbc}) coupled with an exploration mechanism (Line~\ref{alg:line:policy}).

\begin{equation} \label{eq:lbc}
    \mathcal{L}_{BC}(I^s) 
        = \sum_{\left( s_t, s_{t+1} \right) \in I^s} 
        \left | \mathcal{M}_\theta(s_t, s_{t+1}) - \agent(s_t) \right |
\end{equation}
\noindent
The policy then generates new samples ($\mathcal{T}^{\agent}$) that might help $\mathcal{M}$ approximate the unknown ground-truth actions from the expert (Lines~\ref{alg:line:enjoy}-\ref{alg:line:trajectory}).
Given all new samples, \abrev{} generates path signatures $\beta$~\citep{chevyrev2016primer} and uses $\mathcal{D}$ to classify signatures as from the expert or the agent. 
Line~\ref{alg:line:append} updates the discriminator weights with the classification loss in Eq.~\ref{eq:ld},
where $\mathcal{T}_\beta$ are path signatures for all trajectories from expert and agent, $C$ is for the source of the observation (\textit{expert} and \textit{agent}), $y$ is the ground-truth label, and $\hat{y}$ is the source predicted by $\mathcal{D}$.

\begin{equation} \label{eq:ld}
    \mathcal{L}_{\mathcal{D}}(\Tau^\teacher_\beta, \Tau^\agent_\beta) = 
    - \sum_{i=1}^{\mid \Tau_\beta \mid} \sum_{j=1}^{\mid C \mid} y_{ij} log(\hat{y}_{ij}),
\end{equation}

\begin{algorithm}[t]
    \footnotesize
    \caption{\abrev}
    \label{algo:method}
    \begin{algorithmic}[1]
        \State Initialize $\mathcal{M}_\theta$, $\pi_\theta$, and $\mathcal{D}$ with random  weights \label{alg:line:initialize}
        \State $I^s \gets I^{pre}$ s.t. $I^{pre} \gets$ samples from $\pi_\theta$ \label{alg:line:ipre}
        \For { $i \gets 1$ to epochs }
            \State Improve $\mathcal{M}_\theta$ by $\Call{trainM}{I^s}$ \label{alg:line:idm}
            \State Use $\mathcal{M}_\theta$ with $\Tau^\teacher$ to predict $\hat{A}$ \label{alg:line:self}
            \State Improve $\pi_\theta$ by $\text{error}_{\pi_\theta} \gets \Call{behaviourCloning}{\Tau^\teacher, \hat{A}}$ \label{alg:line:policy}
            \State Use $\pi_\theta$ to solve environments $E$ \label{alg:line:enjoy}
            \State $\mathcal{T}^{\pi_\theta} \gets \mathcal{T}^{\pi_\theta} \oplus \left\{(s_0, \hat{a}_0, s_{1}), \cdots, (s_{t-1}, \hat{a}_{t-1}, s_t)\right\}$ \label{alg:line:trajectory}
            \State $I^{pos} \gets I^{pos} \oplus \left\{\forall_{ i \in \mathcal{T}^{\pi_\theta}} \mid \mathcal{D}(\beta(\mathcal{T}^{\pi_\theta}_i)) \textnormal{ is } \top \right\}$ \label{alg:line:append}
            \State $I^s \gets I^{pre} \oplus I^{pos}$ \label{alg:line:concat}
            \If {$\text{error}_{\pi_\theta} \leqslant threshold$} \label{alg:line:early_stop}
            \State Finish training \label{alg:line:break}
            \EndIf
        \EndFor
    \end{algorithmic}
\end{algorithm}

\noindent
Samples classified as expert by $\mathcal{D}$ are added to $I^{pos}$ (Line~\ref{alg:line:append}), which is then combined with the original $I^{pre}$ to form $I^s$ (Line~\ref{alg:line:concat}). 
\abrev{} uses this updated $I^s$ in each iteration for a specified number of epochs~(Line~\ref{alg:line:idm}) or until it no longer improves~(Lines~\ref{alg:line:early_stop}-\ref{alg:line:break}), with an optional hyperparameter $threshold$.

The exploration mechanism allows \abrev{} to deviate from its original action distribution according to the model certainty.
This behaviour is helpful during early iterations when $\mathcal{M}$ is unsure about which action might be responsible for a specific transition.
Since random samples can be very different from the expert's transitions, we can assume that the model does not learn to recognise these transitions and generalises poorly.
Here, we assume that the environment is stochastic, in that multiple actions might occur with a non-zero probability of transition between any pair of states. 
$\mathcal{D}$'s key objective is to discard trajectories that could result in $\mathcal{M}$ getting stuck in bad local minima, and for instance, stop predicting specific actions (underfitting).
Without a discriminator, it would be difficult to ignore signatures that differ considerably from the ground-truth without an environment-specific hyperparameter, hence reducing the method's generalisability.  
Finally, combining these mechanisms makes \abrev{} more sample efficient, allowing for oversampling without misrepresenting the action distributions and overfitting.
Figure~\ref{fig:pipeline} shows the \abrev{} learning cycle in more detail with the different loss functions. 

\subsection{Exploration} \label{sec:sub:exploration}

Exploration is vital for IL methods that use dynamics models to learn how the expert behaves.
It enables policy divergence when the dynamics model is uncertain and increases state diversity, which helps the model approximate labelled transitions from unlabelled ones (expert).
\abrev{} borrows an exploration mechanism from reinforcement learning in continuous domains, in which each action in a policy consists of two outputs: the mean and standard deviation to sample from a Gaussian distribution.
However, unlike traditional reinforcement learning, where a policy receives feedback in the form of the reward function, IL lacks this information. Thus, for a model $\mathbb{M}$ and parameters $\theta$ ($\mathbb{M}_{\theta}$), we employ the sampling mechanism in Eq.~\ref{eq:sampling}, where $\bm{\pi}$ is the usual mathematical constant $3.14\ldots$ and $\varepsilon$, as defined in Eq.~\ref{eq:minkowski}, is used as standard deviation, where $a$ is the ground-truth action (or pseudo-labels from $\mathcal{M}$) and $\hat{a}$ is the action predicted by the model:

\begin{equation} \label{eq:sampling}
    \tilde{a}_{\mathbb{M}_\theta} = \frac{1}{\varepsilon \sqrt{2\bm{\pi}}} e^{-\frac{\left(s^e_t - \mathbb{M}_\theta(S)\right)}{2\varepsilon^2}}
\end{equation}

\begin{equation} \label{eq:minkowski}
    \varepsilon = \left \| a - \hat{a} \right \|^p
\end{equation}

\noindent
In Eq.~\ref{eq:sampling}, $\mathbb{M}$ is either $\mathcal{M}$ or $\pi$, and $\theta$ are the parameters of the model updated for the epoch. Notice that when $p=1$, the model $\mathbb{M}$ uses the absolute value between the predicted and ground-truth labels $\left \| a - \hat{a} \right \|$ and this allows for higher exploration. 

\begin{observation} \label{lemma:exploration}
    If $\mathcal{L}$ is a loss function that monotonically decreases a model's $\mathbb{M}$ error as it approximates the ground-truth function, eventually $\left \| a - \hat{a} \right \| <1$. 
    If we then use $p > 1$ in Eq.~\ref{eq:minkowski}, $\varepsilon$ will exponentially decrease.
\end{observation}

Given all of the above, Eq.~\ref{eq:minkowski} offers a trade-off between exploration and exploitation. 
Since $\varepsilon$ is the standard deviation for the exploration function, as the model's predictions get closer to the ground-truth and pseudo-labels, the clusters will have lower variance because the exploration ratio is directly correlated to the model's error.

In Alg.~\ref{algo:method}, functions $\Call{trainM}{}$ (ln.~\ref{alg:line:idm}) and $\Call{behaviourCloning}{}$ (ln.~\ref{alg:line:policy}) use this adaptation to adjust the exploration ratio depending on how close the model's predictions are to the ground-truth (or pseudo-labels), in accordance with the standard deviation of the Gaussian distribution.
This mechanism also has the benefit of not having to predict information beyond the agent's actions, such as standard deviation, instead obtaining this directly from the model's error.
For deterministic behaviour, we can assume that the standard deviation for the model is $0$ and use the model's output since sampling from a Gaussian distribution with average $x$ and deviation $0$ equals $x$. 

\subsection{Goal-aware function} \label{sec:sub:goal}

Developing a goal-aware function may not be a trivial task.
For environments with a well-defined goal, such as CartPole~\citep{barto1983neuronlike}, which defines the goal to be balancing the pole for $195$ steps, a goal-identification function could simply classify all trajectories that reach $195$ steps as optimal.
In this work, we formally define trajectories as:
\begin{definition}
    A trajectory $\tau$ is a finite sequence of states $(s_1,\ldots,s_n)$ where for each $1 \leq i < n$, $s_{i+1}$ is obtained from $s_i$ via the execution of some action. We use the term $(\tau_t^1,\tau_t^2,\ldots,\tau_t^d) \in \mathbb{R}^d$ to denote the particular state $s_t$  ($1\leqslant t \leqslant n$) within the trajectory $\tau$.
\end{definition}

\noindent
However, recall that in the context of IL, the agent has no access to the reward signal, and as environments grow in complexity, such a function becomes even harder to encode. 
By contrast, some environments have no prescribed goal. 
For example, the Ant environment requires the agent to walk as far as possible without falling, but with no defined cap on the number of time steps~\cite{schulman2015high}.
Thus, existing IL approaches~\cite{monteiro2020augmented,gavenski2020imitating,gavenski2021self} often rely on manually defined goal-aware functions, which have the benefit of dispensing with the alignment of the environment's goal.
For example, we might define a specific trajectory as required in the Ant environment. 
Unless the agent reaches all points in this trajectory, our goal-aware function does not classify the episode as successful.
However, this creates a degree of unwanted complexity in a learning algorithm and a cumbersome process as the number of environments grows.
Yet, trajectories may carry relevant information for \abrev{} since they approximate $I^s$'s samples from $\Tau^\teacher$~\cite{gavenski2020imitating}.
Therefore, \abrev{} tries to classify trajectories that are close to $\Tau^\teacher$ instead of successful ones.

Nevertheless, identifying whether samples are near $\Tau^\teacher$ is also difficult.
If we consider a stationary agent, we might discard samples that allow $\mathcal{M}$ to better predict transitions due to their distance to the $\teacher$ states alone.
But, if we consider whole trajectories, it might be difficult to identify middling trajectories needed to close the gap between $\Tau^\agent$ and $\Tau^\teacher$, and better generalise~\cite{Gavenski2022how}.
Therefore, \abrev{} needs a function that (i) simplifies comparisons between trajectories and (ii) allows $\mathcal{M}$ to receive suboptimal samples.

For the first problem, previous work~\cite{pavse2020ridm} dealt with the issue of trajectory length by using the average of all states up to a point in time to account for the trajectory changes.
Conversely, we use path signatures~\cite{chevyrev2016primer}, which are fixed-length feature vectors that are used to represent multi-dimensional time series (i.e., trajectories).
A path signature is computed by the function $\beta$ comprehensively defined in Section~3 of \citet{yang2022developing}, succinctly summarised in the definition below (see Supplementary Material for more detail).\footnote{In our experiments, $\beta$  (Line~\ref{alg:line:append}, Algorithm~\ref{algo:method}) was computed using the implementation provided by \cite{kidger2021signatory}.}

\begin{definition} \label{def:signatutre}
    Let a trajectory $\tau$ of a countable length between $[1, n]$ ($n \in \mathbb{N}$), where each state is a vector in $\mathbb{R}^d$ with dimensions indexed by a collection of indices $i_1, \cdots, i_k \in \left \{ 1, \cdots, d \right \}$.
    Let the recursively computed path signature $\beta$ for a trajectory $\tau$ for any $k \geqslant 1$ and time $t$ ($1 \leqslant t \leqslant n$) be:
    \begin{equation} \label{eq:recursive_signature}
        \beta(\tau)^{i_1,\cdots,i_k}_{1, t} =
        \int_{1 < s \leqslant t} \beta(\tau)^{i_1, \cdots, i_{k-1}}_{1, s} d\tau^{i_k}_s.
    \end{equation}
    Then, the signature of a trajectory $\tau:[1, n] \rightarrow \mathbb{R}^d$ is the collection of all the iterated integrals of $\tau$:
    \begin{equation}
        \begin{aligned}
            \beta(\tau)_{1,n}^{1,\ldots, i_k} = \bigg(
                & 1, \beta(\tau)^1_{1,n}, \cdots, \beta(\tau)^d_{1,n}, \beta(\tau)^{1,1}_{1,n}, \cdots, \\
                & \beta(\tau)^{1,d}_{1,n}, \beta(\tau)^{2,1}_{1,n}, \ldots, \beta(\tau)^{i_1, i_2, \cdots, i_k}_{1,n} 
            \bigg),
        \end{aligned}
    \end{equation}
    where the zero-th term is conventionally equal to 1, and $k$ is defined as the $k$-th level of the signature, which defines the finite collection of all terms $\beta(\tau)^{i_1, \cdots, i_k}_{1,n}$ for the multi-index of length $k$. For example, when $k = d$, the last term would be $\beta(\tau)_{1, n}^{d, d, \cdots, d}$.
\end{definition}

Path signatures allow \abrev{} to solve the issue of comparing two trajectories and encoding different characteristics that may be relevant when classifying how close a new trajectory is from $\Tau^\teacher$.
By using path signatures generated from trajectories in $\Tau^\agent$ and $\Tau^\teacher$, \abrev{} benefits from:
\begin{enumerate*}[label=(\roman*)]
    \item a common signature size, regardless of the original length of trajectories, helping the discriminator not to discriminate against longer trajectories;
    \item independence of environment characteristics embedded in the data (avoiding the need for re-parametrisation for each environment); and
    \item the preservation of the uniqueness of trajectories via the non-linearity of the signatures.
\end{enumerate*}

The use of signatures still requires some manual intervention in \abrev{} to define how close a trajectory needs to be before adding it to $I^s$ (i.e., an appropriate similarity threshold).
To prevent the need for manually defining this threshold, \abrev{} uses a discriminator model $\mathcal{D}$ to discriminate between $\agent$ and~$\teacher$ trajectories, which optimises Eq.~\ref{eq:ld}.
This yields a non-greedy sampling mechanism by using a model to classify expert and non-expert trajectories.

In summary, \abrev{}'s goal-aware function works by computing a signature $\beta(\tau)$ of a trajectory $\tau$ and feeds it into the discriminator model $\mathcal{D}$, which classifies whether the source of the trajectory is $\agent$ or $\teacher$.
If $\mathcal{D}$ classifies the source of an agent's trajectory as the expert, then \abrev{} appends the trajectory into $I^s$, helping $\mathcal{M}$ better understand how the transition function $T$ works in the environment.

\begin{table*}[tp]
    \scriptsize
    \centering
    \caption{\abrev{} and baselines AER and $\mathcal{P}$ results for all environments. All results are the average of $50$ trajectories.}
    \label{tab:results}
    
    \begin{tabular*}{\textwidth}{@{\extracolsep{\fill}}llrrrrr}
        \toprule
        Algorithm & Metric & Ant & Pendulum & Swimmer & Hopper & HalfCheetah \\
        \midrule 
        \midrule
        \multirow{1}{*}{Random} & AER & $-65.11 \pm 106.16$ & $5.70 \pm 3.26$ & $0.73 \pm 11.44$ & $17.92 \pm 16.02$ & $-293.13 \pm 82.12$ \\
                               & $\mathcal{P}$ & $0$ & $0$ & $0$ & $0$ & $0$ \\
        \midrule 
        \multirow{1}{*}{Expert} & AER & $5544.65 \pm 76.11$ & $1000 \pm 0$ & $259.52 \pm 1.92$ & $3589.88 \pm 2.43$ & $7561.78 \pm 181.41$ \\
                               & $\mathcal{P}$ & $1$ & $1$ & $1$ & $1$ & $1$ \\
        \midrule 
        \midrule
        \multirow{1}{*}{\abrev} & AER & $\mathbf{6091 \pm 801.2}$ & $\mathbf{1000 \pm 0}$ & $\mathbf{334.6 \pm 3.45}$ & $\mathbf{3589 \pm 178.2}$ & $\mathbf{7100.6434 \pm 90.1775}$ \\
                                & $\mathcal{P}$ & $\mathbf{1.0974 \pm 0.1372}$ & $\mathbf{1 \pm 0}$ & $\mathbf{1.2901 \pm 0.0128}$ & $\mathbf{0.9998 \pm 0.0487}$ & $\mathbf{0.9413 \pm 0.0115}$ \\
        \midrule
        \multirow{1}{*}{OPOLO}  
        & AER & $5508.6807 \pm 930.7590$ & $\mathbf{1000 \pm 0}$ & $253.3297 \pm 3.4771$ & $3428.6405 \pm 420.3285$ & $7004.65 \pm 568.66$ \\
        & $\mathcal{P}$ & $0.9935 \pm 0.1659$ & $\mathbf{1 \pm 0}$ & $0.9761 \pm 0.0134$ & $0.9549 \pm 0.1177$ & $0.9291 \pm 0.0724$  \\
        \midrule
        \multirow{1}{*}{MobILE} & AER & $995.5 \pm 25.65$ & $111.7 \pm 31.25$ & $130.7 \pm 24.36$ & $2035 \pm 192.95$ & $4721.5 \pm 364.5$ \\
                               & $\mathcal{P}$ & $0.1891 \pm 0.0047$ & $0.1066 \pm 0.0313$ & $0.5022 \pm 0.0968$ & $0.5647 \pm 0.0531$ & $0.5647 \pm 0.0454$ \\
        \midrule
        \multirow{1}{*}{BCO}    & AER & $1529 \pm 980.86$ & $521 \pm 178.9$ & $257.38 \pm 4.28$ & $1845.66 \pm 628.41$ & $3881.10 \pm 938.81$ \\
                               & $\mathcal{P}$ & $0.2842 \pm 0.1724$ & $0.5675 \pm 0.1785$ & $0.9917 \pm 0.0166$ & $0.5177 \pm 0.1765$ & $0.5117 \pm 0.1217$ \\
        \bottomrule
    \end{tabular*}
\end{table*}

\subsection{Sample efficiency} \label{sec:sub:sample}

Besides approximating the expert policy, IL methods focus on efficiently using expert samples.
This focus happens since expert samples are hard to obtain.
Thus, creating more efficient methods, \ie, that require fewer samples, allows for more useable approaches.
Some recent strategies~\cite{kidambi2021mobile,zhu2020off} minimise the number of required samples but depend on strong assumptions (see Section~\ref{sec:sub:results}) or manual intervention for each new environment. 
For comparison, \abrev{} uses $10$ expert episodes -- a number similar to~\citet{zhu2020off} and~\citet{kidambi2021mobile}, but without requiring manual intervention for each environment.
\abrev{} relies on up-scaling $\mathcal{T}^\teacher$ to increase the number of observations $\agent$ sees before interacting with the environment.
Although trivial, this strategy works because \abrev{} is self-supervised and has an exploration mechanism.
This strategy helps in two ways:
\begin{enumerate*}[label=(\roman*)]
    \item for each epoch all pseudo-labels differ in all transitions due to the exploration mechanism (Line~\ref{alg:line:idm}, Algorithm~\ref{algo:method}); and
    \item increasing the number of samples $\agent$ receives allows for more updates before sampling new experiences from the environment.
\end{enumerate*}
By applying its exploration mechanism to each observation individually and sampling exploration values from a distribution, \abrev{} ensures that each observation has unique action values, reducing the risk of misrepresenting the ground-truth action distribution.

\section{Experimental Results} \label{sec:experimental}

We compared \abrev{}'s results against three key related methods. 
Behavioral Cloning from Observations (BCO)~\citep{TorabiEtAl2018}, which is usually used as a baseline, and two of the most efficient LfO methods:  Off-Policy Imitation Learning from Observations (OPOLO)~\citep{zhu2020off}, and Model-Based Imitation Learning From
Observation Alone (MobILE)~\citep{kidambi2021mobile}. 
We experimented with five commonly used environments: Ant, Half Cheetah, Hopper, Swimmer, and Pendulum.\footnote{The Supplementary Material briefly describes these environments and the neural networks topology.}
Each method was run for $50$ episodes, with the environment reset when the agent falls or after $1,000$ steps.
Each episode was run using random seeds to test the agent's ability to generalise.

\subsection{Implementation and Metrics} \label{sec:sub:metrics}

We used PyTorch to implement our agent and optimise the loss functions in Eq.~\ref{eq:lm}-\ref{eq:ld} via Adam~\cite{kingma2014adam} and Imitation Datasets~\cite{gavenski2024imitation} to collect the expert data.
As for the exploration mechanism in Eq.~\ref{eq:minkowski}, we use $p=1$ for $\varepsilon$ due to all environments actions being in the interval $[-1, 1]$, and using $p > 1$ would significantly diminish the gap between predicted and ground-truth actions (as defined in Definition~\ref{lemma:exploration}). 
In the supplementary material, we provide all learning rates and discuss hyperparameter sensitivity in more detail, but we note that \abrev{} is not very sensitive to precise hyperparameters. 

We evaluated all approaches using the \textit{Average Episodic Reward} (AER) metric (Eq.~\ref{eq:aer}) and use \textit{Performance} ($\mathcal{P}$) (Eq.~\ref{eq:perf}).
AER is the average accumulated reward for a policy $\policy$ over $n$ number of episodes in $t$ number of steps:

\begin{equation} \label{eq:aer}
    AER(\policy) = \frac{1}{n}\sum^{n}_{i=1} \sum_{j=1}^{t} \gamma^t r(s_{ij}, \policy(s_{ij})).
\end{equation}

\noindent
On the other hand, $\mathcal{P}$ normalises between random and expert policies rewards, where performance $0$ corresponds to random policy $\random$ performance, and $1$ is for expert policy $\teacher$ performance.

\begin{equation} \label{eq:perf}
    \mathcal{P}_\tau(\policy) = \frac{AER(\policy) - AER(\random)}{AER(\teacher) - AER(\random)}
\end{equation}

\noindent
Note that a negative value for $\mathcal{P}$ indicates a reward for the agent lower than a random agent's and a value higher than $1$ indicates that the agent's reward is higher than the expert's.
All results in Table~\ref{tab:results} are the average and standard deviation in five different experiments.
We do not report accuracy since achieving high accuracy does not necessarily translate into a high reward for the agent.

\subsection{Results} \label{sec:sub:results}

We trained all methods using $10$ expert trajectories. 
Table~\ref{tab:results} shows how each method performed in the five environments.
\abrev{} had the best overall results in all environments.
It consistently achieved results similar to the expert, surpassing it on Ant and Swimmer and achieving the maximum reward for the Pendulum environment.
\abrev's performance was close to the expert's in Hopper but a little lower in HalfCheetah -- likely due to the higher standard deviation from the ground-truth actions in both environments.
In the Swimmer environment, BCO and OPOLO achieved AER and performance similar to the expert, while \abrev{} outperformed it by $0.29$ points.
The same happened in the Ant environment, where \abrev{} surpassed the expert by $\approx 484$ reward points.
We hypothesise this is due to \abrev{}'s explorative nature and its ability to acquire new samples that the discriminator judges to come from the expert. 

Comparing \abrev{} to other methods, we see that OPOLO had the closest performance to \abrev's in almost all environments. 
We attribute \abrev{}'s better performance than OPOLO's due to the fact that OPOLO's problem formulation assumes that the environment follows an injective MDP, which cannot be guaranteed with random seeds.
For this work, we believe that it is more important for an agent to be able to correct its initial states into a successful trajectory than to be optimal in a single setting.
Moreover, we notice that for the Pendulum environment, OPOLO only achieved the optimal reward when clipping the actions between $[-1, 1]$, which \abrev{} does not require.
When the actions are not clipped, OPOLO accumulates $\approx 9.55$ reward points, a performance similar to the random policy.
We opted not to clip \abrev{}'s actions, so that the method would not require any previous environment knowledge.

BCO requires more expert trajectories to achieve better results.
In its original work, BCO used $5 \times 10^5$ samples for $\mathcal{M}$ and more than $1,000$ expert trajectories for its policy, which may be unrealistic for many domains. 
Nevertheless, BCO achieved almost expert results in the Swimmer environment and higher rewards than MobILE in almost all other environments, with the exception of HalfCheetah.
We believe BCO outperforms MobILE because the latter assumes that each environment has a fixed initial state, which does not happen since the gym suite alters each initial state according to some parametrised intervals and its current seed.

As for MobILE, we used the same number of trajectories as in its original work.
We observe that MobILE suffers from three different issues:
\begin{enumerate*}[label=(\roman*)]
    \item it is ensemble;
    \item has domain knowledge embedded into the algorithm (not publicly available); and 
    \item its results are difficult to reproduce, because of the large number of hyperparameters on which they depend.
\end{enumerate*}
During our experimentation, we observed that some approaches underperform when using an expert with strict movement constraints. 
To some extent, when obtaining $\mathcal{T}^\teacher$, all environments are susceptible to this, but MoBILE was especially impacted.
We believe that this strict movement pattern is difficult for all methods to learn since the impact of the variations cannot be immediately perceived.
The lack of reproducibility is a major drawback of MobILE, from which \abrev{} does not suffer.
By using path signatures, which is a non-parametric encoding technique, \abrev{} is left with only two different parameters: the network size and the learning rate.
We used \citeauthor{smith2017cyclical}'s work~(\citeyear{smith2017cyclical}) as a guide for finding optimal values for these parameters.\footnote{We followed the original network topology for a fair comparison.} 

Finally, in environments with broadly distributed expert actions like Ant, Pendulum, and Hopper, \abrev{} matches expert performance in fewer iterations than the other methods.
However, in environments where actions are more concentrated (Swimmer and HalfCheetah), \abrev{} takes longer to match the expert.

\begin{figure*}[!t]
    \centering
    \subfloat[$\pi_\theta$ trained with a scheduler.\label{fig:scheduler}]{\includegraphics[width=0.25\textwidth]{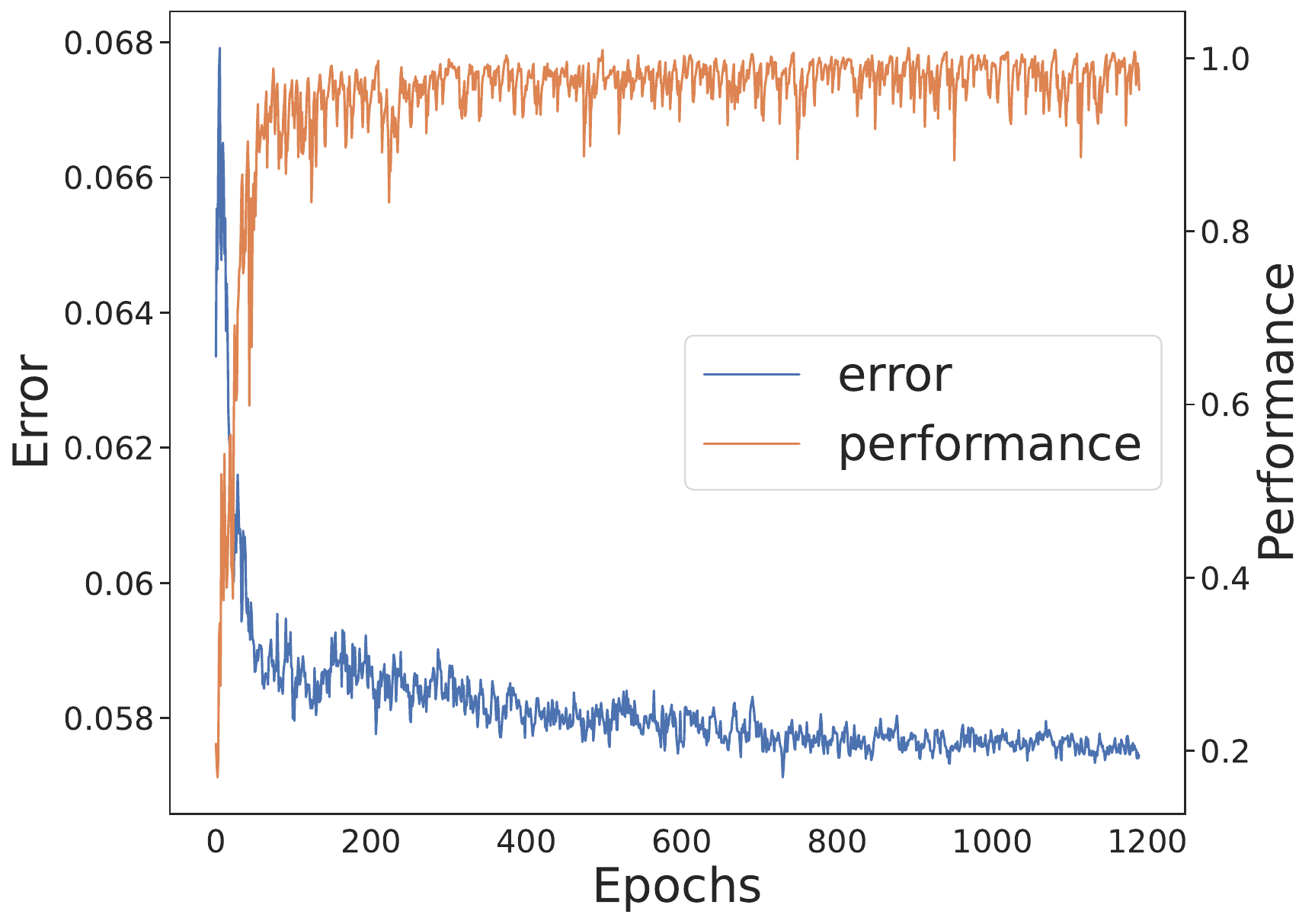}}
    \hfill
    \subfloat[$\pi_\theta^*$ trained with no scheduler.\label{fig:noscheduler}]{\includegraphics[width=0.25\textwidth]{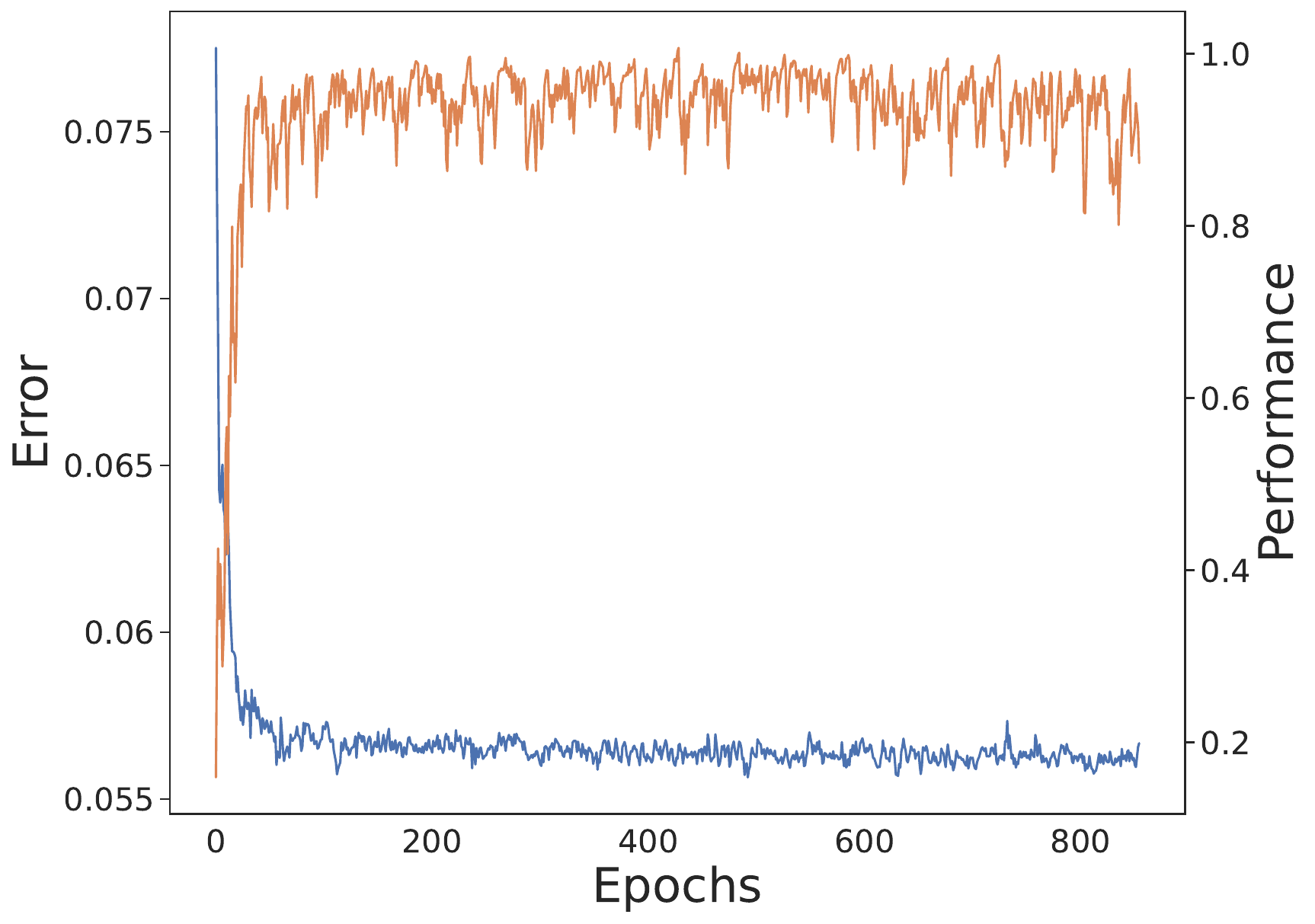}}
    \hfill
    \subfloat[Signature difference in Ant-v3.\label{fig:signature}]{\includegraphics[width=0.25\textwidth]{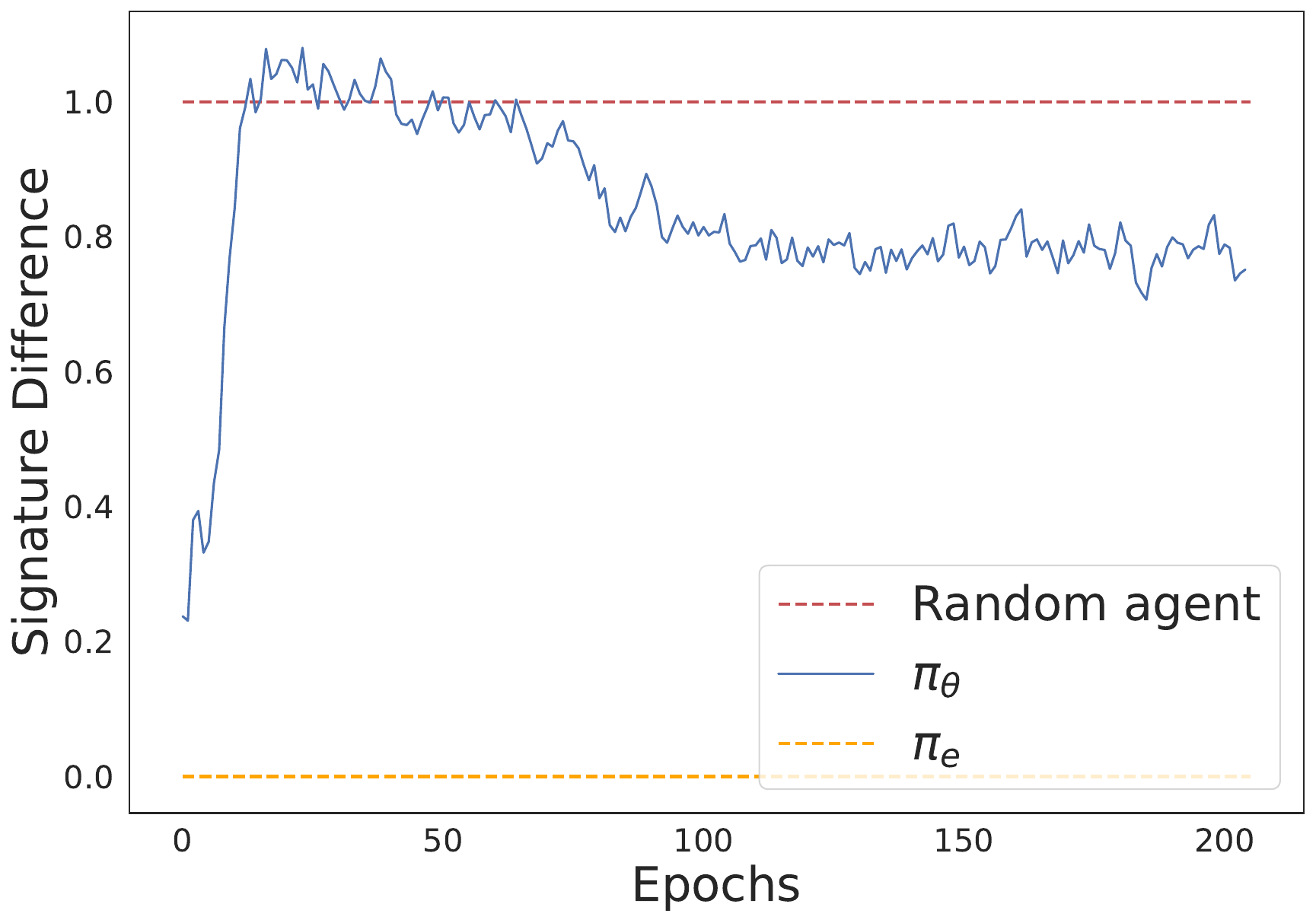}
}
    \caption{(a) and (b) show ground-truth error for $\mathcal{M}$ and $\mathcal{P}$. (c) shows the normalised difference between $\pi_e$, $\pi_\theta$ and random signatures: $0$ is equivalent to expert, and $\geqslant 1$ means equal or worse than random policy signature.}
    \label{fig:ablation}
\end{figure*}

\section{Discussion} \label{sec:discussion}

In this section, we consider some key aspects of \abrev's behaviour:
\begin{enumerate*}[label=(\roman*)]
    \item how \abrev{} learns with different sample amounts; 
    \item how it approximates predictions to the ground-truth actions of the expert;
    \item how similar each signature becomes to all trajectories over time;
    \item how different action distributions affect \abrev; and
    \item how $I^s$ behaves over time.
\end{enumerate*}

\subsection{Sample Efficiency} \label{sec:sub:ab:samples}

In order to understand \abrev's sample efficiency, we experimented with three different amounts of expert episodes in the Ant environment.
Ant provides an ideal setting due to its balanced learning complexity and shorter training times. 
Table~\ref{tab:ab:efficiency} shows the AER and $\mathcal{P}$ results using $1$, $10$, and $100$ trajectories.
As expected, \abrev{} does not achieve good results when using a single trajectory.
This is because $\agent$ has no information regarding different initialisation and trajectory deviations.
This behaviour is intrinsic to behavioural cloning where, without sufficient information, the policy tends not to generalise~\citep{le2018imitation}.

\begin{table}[b]
    \footnotesize
    \centering
    \caption{\abrev{}'s AER and $\mathcal{P}$ values for different Ant dataset sizes.}
    \label{tab:ab:efficiency}
    \begin{tabular*}{\columnwidth}{c@{\extracolsep{\fill}}rr}
        \toprule
        Trajectories & \multicolumn{1}{c}{AER} & \multicolumn{1}{c}{$\mathcal{P}$} \\
        \midrule
        1 & $1003 \pm 1999$ & 0.18 \\
        10 & $\mathbf{6091 \pm 801.2}$ & $\mathbf{1.1}$ \\
        100 & $6026 \pm 725.86$ & 1.09 \\
        \bottomrule
    \end{tabular*}
\end{table}

Interestingly, \abrev{} achieves $65$ \emph{fewer} reward points when using $100$ trajectories than when it uses $10$.
We attribute this to the following:
\begin{enumerate*}[label=(\roman*)]
    \item when used in a LfO scenario, BC methods usually fail to scale according to the number of samples due to \textit{compounding error}~\citep{swamy2021moments}; and
    \item increasing the number of expert samples decreases the deviation from $\teacher$ trajectories, resulting in overfitting and a worse $\agent$. 
\end{enumerate*}
Since it achieves expert results for almost all environments, we do not consider this behaviour a limitation of \abrev.
Nevertheless, we hypothesise that using different strategies might result in an increase in performance when its data pool is increased.
We also hypothesise that using incomplete or faulty trajectories might help \abrev{} since it would not have so much data for all points in a trajectory, reducing overfit. 
Fine-tuning the exact number of expert trajectories requires some experimentation for each environment.

\subsection{Ground-truth error over time} \label{sec:sub:ab:gt}

A concern for self-supervised IL methods is how to approximate pseudo-labels to ground-truth actions from the expert.
However, approximating $I^s$ samples to those from the experts is not always best.
There might be samples that are between the experts' and $I^{pre}$ that can help $\mathcal{M}$ smoothly close the gap between equally distributed and the ground-truth distribution~\citep{gavenski2021self}.
Since \abrev{} uses exploration to learn and this exploration mechanism relies on $\mathcal{M}$'s error, achieving lower error margins early might lead to less exploration and poorer results.
It would therefore be better to have a consistent stream of new samples, to maintain the error marginally high but not a significant number of new samples since this could keep $\mathcal{M}$'s error too high or even collapse the network, \ie, updating all weights drastically and requiring a higher learning rate.

Table~\ref{tab:gt} shows that using two different procedures and achieving two different policies with different error margins and weights yields similar results for the error margins in both methods but not performance and AER.
Figure~\ref{fig:scheduler} shows the learning results (error and performance) for $\agent$, which uses weight decay and a scheduler during training, and Figure~\ref{fig:noscheduler} shows $\pi^*_\theta$, with no weight decay and a learning rate scheduler.
We observe that both policies achieve a similarly consistent error margin.
However, when comparing the average performances in a single episode, $\agent$ achieves $29$ more reward points with a lower variation. 
While using different strategies for classifying the action might help \abrev{} with this behaviour, we use a similar topology to the one used by the models compared.

\begin{table}[ht]
    \footnotesize
    \centering
    \caption{$\mathcal{M}$'s ground-truth error and policy's AER and~$\mathcal{P}$ for Ant.}
    \label{tab:gt}
    \begin{tabular*}{\columnwidth}{c@{\extracolsep{\fill}}ccc}
        \toprule
            Method & Error & AER & $\mathcal{P}$ \\
        \midrule
            $\pi_\theta$ & $\mathbf{0.0571}$ & $\mathbf{5610}$ & $\mathbf{1.0116}$ \\
            $\pi_\theta^*$ & 0.0556 & 5581 & 1.0065 \\
        \bottomrule
    \end{tabular*}
\end{table}

\subsection{Signature approximation over time} \label{sec:sub:signature}

\begin{figure*}[t]
    \subfloat{\includegraphics[width=0.1\textwidth]{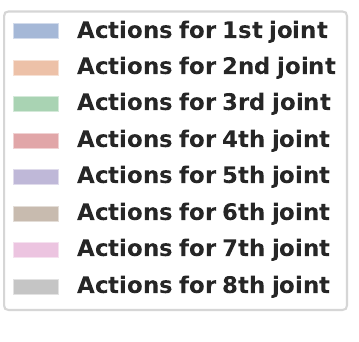}}
    \hfill
    \addtocounter{subfigure}{-1}
    \subfloat[Ant~($\teacher$)\label{fig:ant:expert}]{\includegraphics[width=0.2\textwidth]{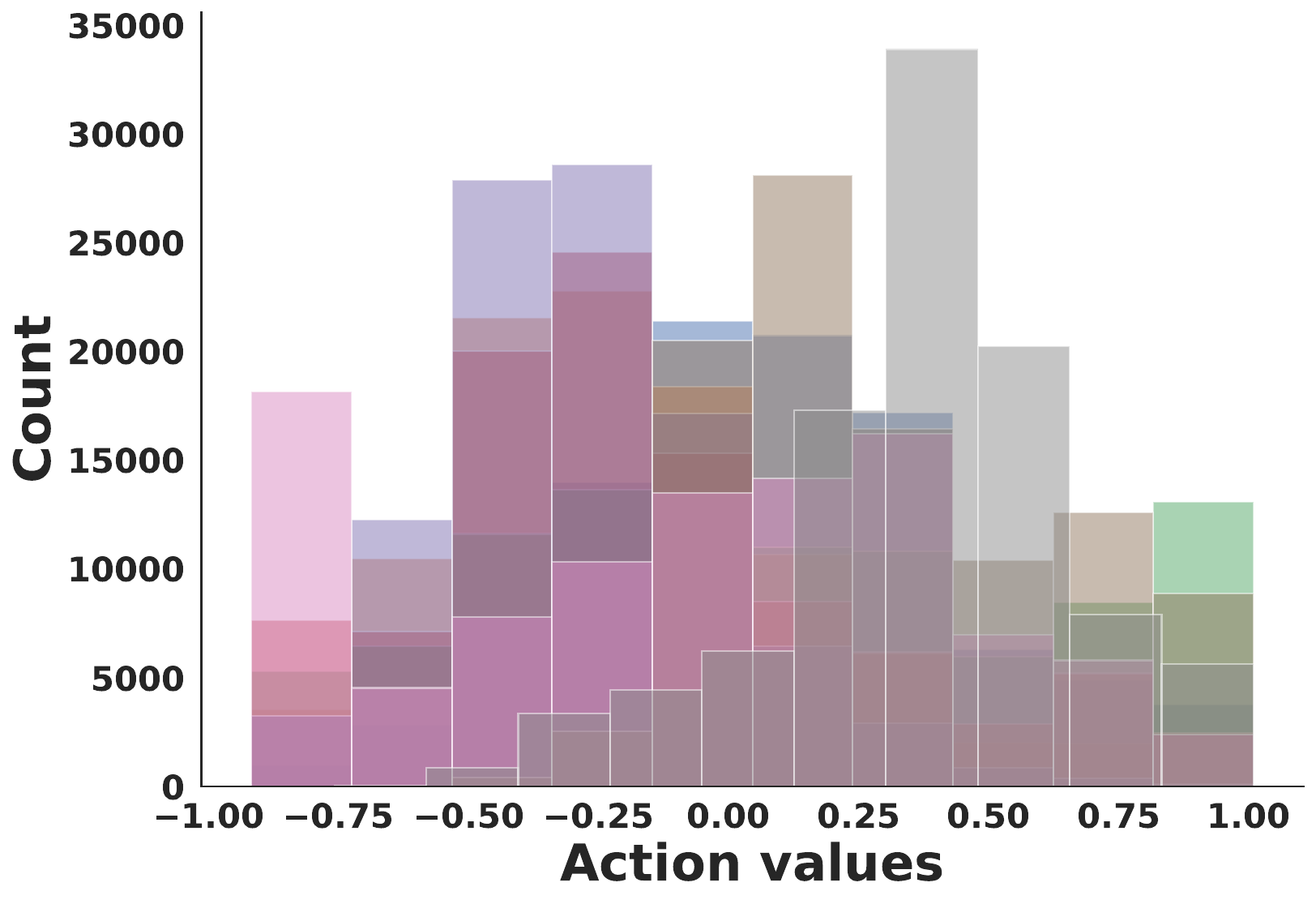}}
    \hfill
    \subfloat[Ant~($\agent$)\label{fig:ant:policy}]{\includegraphics[width=0.2\textwidth]{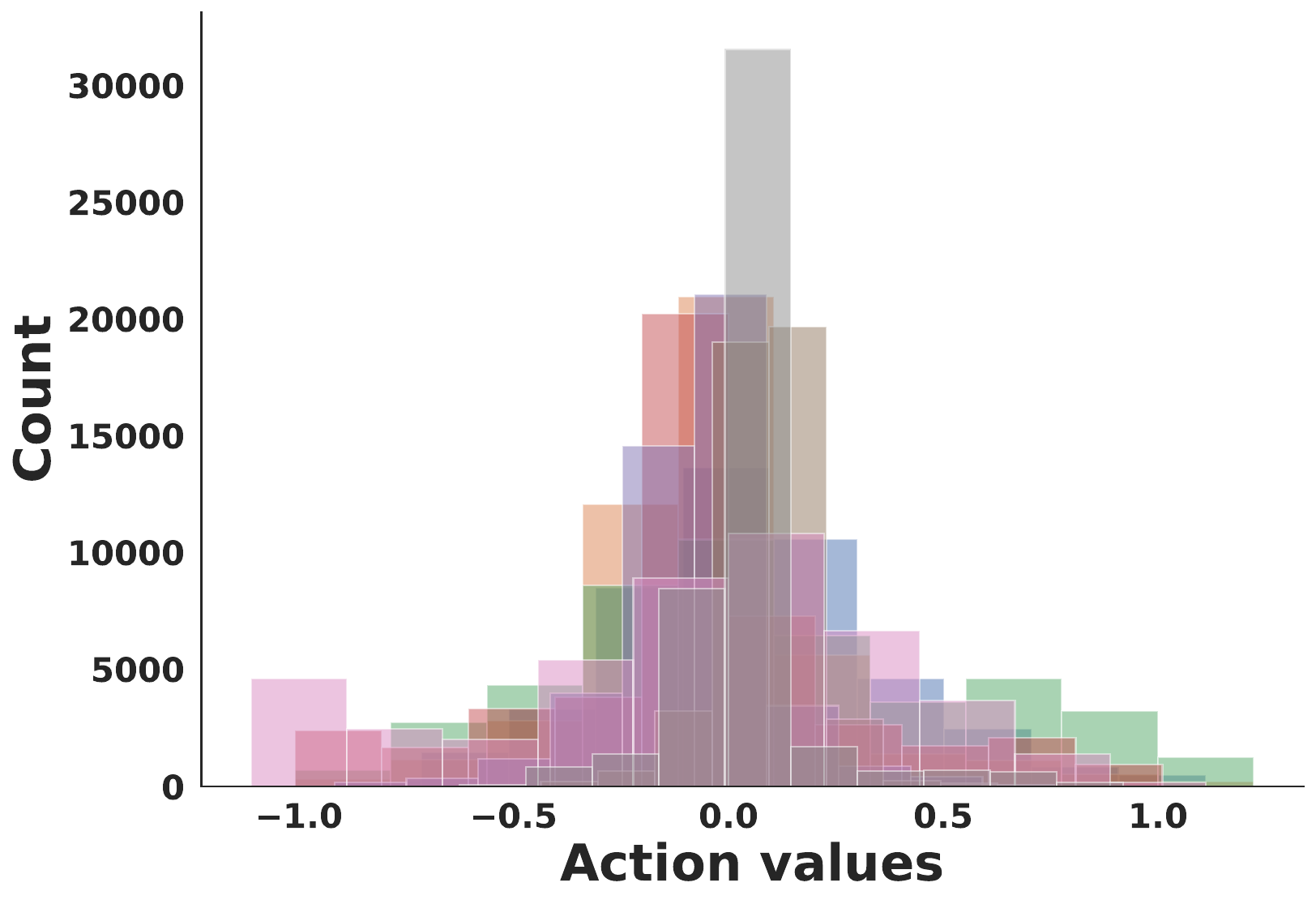}}
    \hfill
    \subfloat[HalfCheetah~($\teacher$)\label{fig:cheetah:expert}]{\includegraphics[width=0.2\textwidth]{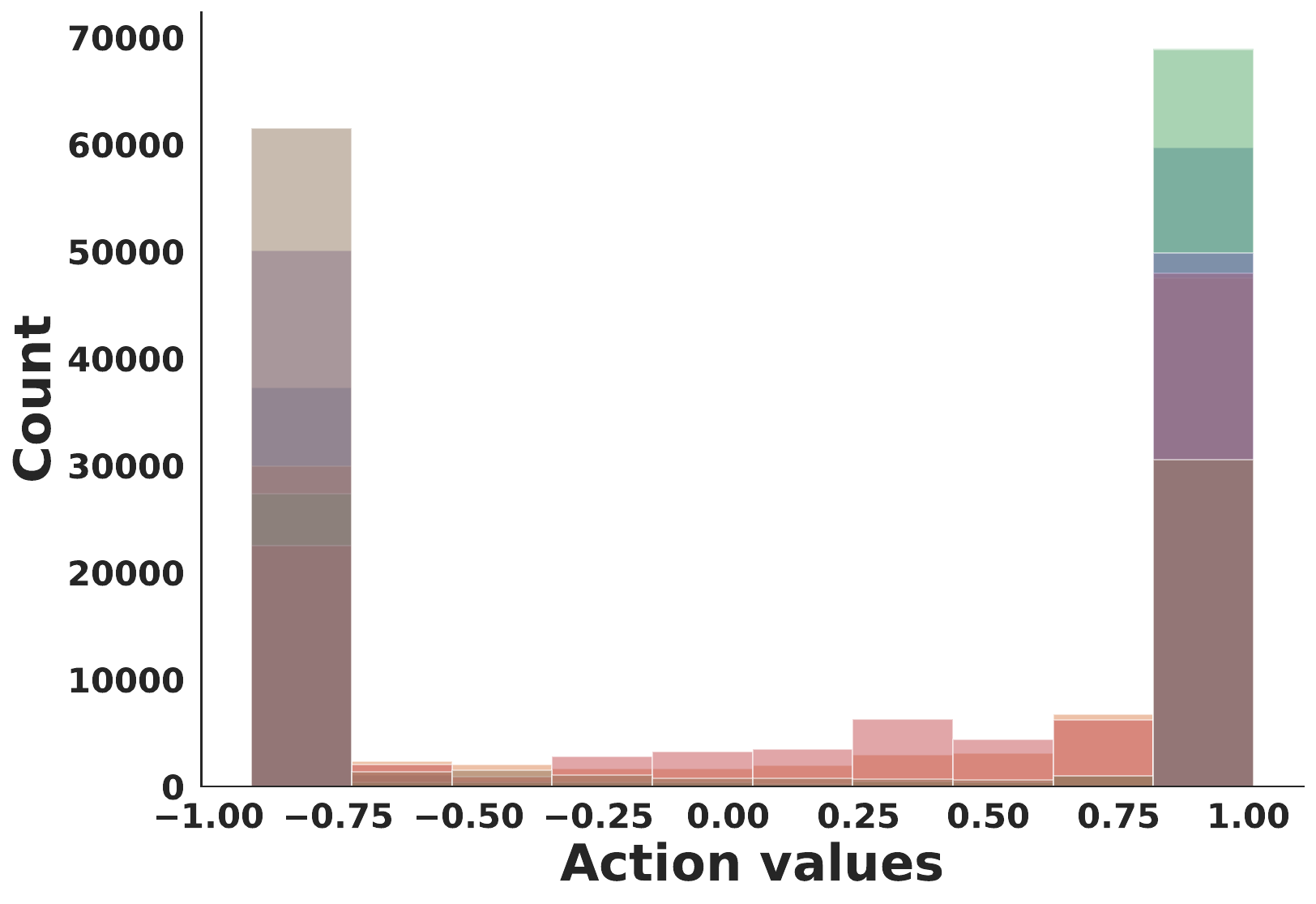}}
    \hfill
    \subfloat[HalfCheetah~($\agent$)\label{fig:cheetah:policy}]{\includegraphics[width=0.2\textwidth]{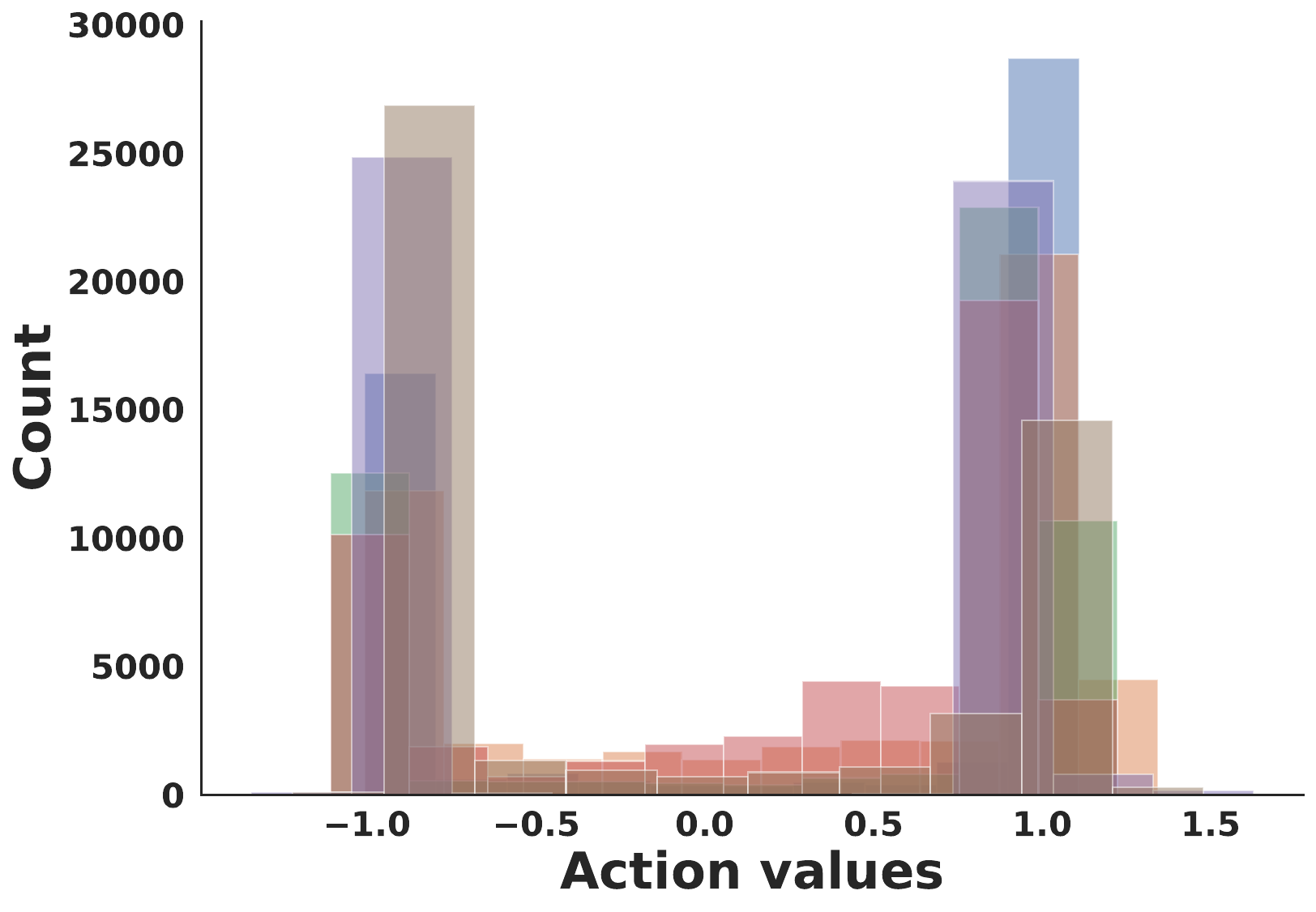}}
    \caption{Distribution of expert actions for Ant and HalfCheetah environments.}
    \label{fig:distributions}
\end{figure*}

Given Definition~\ref{def:signatutre}, trajectories that are similar should be closer in the feature space, while those that do not share any states should be farther apart.
Figure~\ref{fig:signature} shows the Manhattan distance between $\agent$ and $\teacher$ trajectories during the first $200$ iterations.
The difference is normalised between trajectories from random and expert agents.
Hence, a difference greater than $1$ means that the agent's signature path is farther from $\teacher$ than a random agent's.
As expected, during early iterations, \abrev{} produces episodes that are farther than the random agent since $\mathcal{M}$ has to learn state transitions before $\agent$ can learn how to behave in the environment.
We see similar behaviour from the discriminator $\mathcal{D}$.
In the initial iterations, it allows multiple trajectories to be appended to $I^s$ due to its poor performance in discriminating between generated and expert trajectories. Once $\mathcal{D}$ learns to classify correctly, it is only `fooled' by $\approx 18\%$ of trajectories.

As $\agent$ increases its performance, the distance between $\agent$ and $\teacher$ signatures decreases.
Similarly, $\mathcal{D}$ has a harder time distinguishing from expert and $\agent$.
We observe that $\mathcal{D}$'s results are as expected.
By allowing these early trajectories to append into $I^s$, which had not achieved any goals, it allows $\mathcal{M}$ to learn from samples outside its randomly distributed ones.
Since it only allows a few samples, $\mathcal{M}$ does not stop to predict actions due to skewed samples.
But as $\mathcal{D}$ improves its classification performance, it forces $\agent$ indirectly to be closer to the expert behaviour, therefore, achieving higher rewards.
Using the gradient signal from $\mathcal{D}$ is likely to improve $\agent$'s performance further, but this adaptation would require the policy also to predict the next state, \eg, in a mechanism similar to the one used in \citeauthor{EdwardsEtAl2019}'s work~\cite{EdwardsEtAl2019}.

\subsection{Effects of Gaussian exploration} \label{sec:sub:distribution}

Since we observed that \abrev{} has a different behaviour for environments with different action distributions, we analyse Ant and HalfCheetah to understand the disparities between $\agent$ and $\teacher$ action predictions.
Figure~\ref{fig:distributions} displays all distributions for $50$ trajectories from the expert and trained policies.
Note that for these actions, $\agent$ is not using its exploration mechanism, that is, the policy is greedy.
In all environments, the distribution from $\agent$ actions differs from the expert ones.
However, we observe that $\agent$ actions have a higher intra-cluster variance than the expert ones.
We believe this behaviour is due to \abrev{}'s exploration mechanism sampling from a Gaussian distribution, making it learn to have a higher variance around the average of an action (considering the error rate from Table~\ref{tab:gt}).
Therefore, the exploration mechanism makes it difficult to approximate distributions that do not follow this pattern, such as HalfCheetah.

We also note that \abrev{} has more difficulty achieving better results in environments with sparse action distributions.
If we compare Figures~\ref{fig:cheetah:expert} and~\ref{fig:cheetah:policy}, it is evident that \abrev{} achieves actions near both limits, \ie, $-1$ and $1$; however, it has a harder time predicting actions near the limit.
In contrast, although both distributions from Figures~\ref{fig:ant:expert} and~\ref{fig:ant:policy} are unequal, we observe a more concentrated action cluster around $0$, which helps $\agent$ achieve better results.
We see this behaviour as a limitation of \abrev{} since selecting a new sampling distribution would require knowing beforehand how an expert behaves.
However, we also hypothesise that training for a period without exploration and fine-tuning $\agent$ with $\mathcal{M}$'s pseudo-labels would minimise this impact.
Further training $\agent$ with no exploration, we observe an increase in all environments, although not significantly.

\subsection{$I^s$ size over time}

\begin{figure}[b]
    \centering
    \includegraphics[width=0.75\columnwidth]{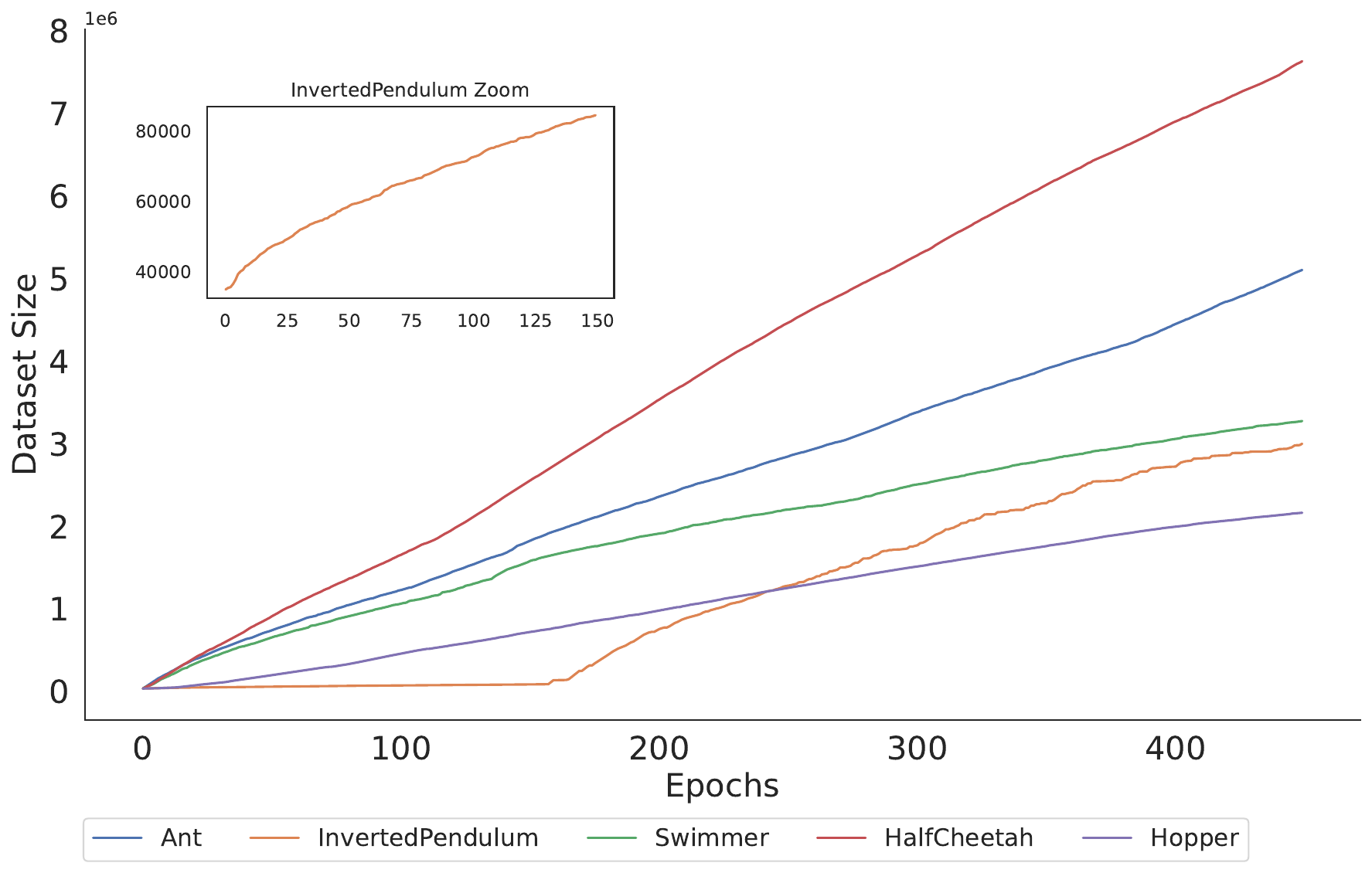}
    \caption{Size of $I^s \times$ epochs for all environments.}
    \label{fig:idm_size}
\end{figure}

The use of the discriminator $\mathcal{D}$ allows \abrev{} to start with fewer random samples since it appends samples on almost every iteration.
However, increasing $I^s$ on each epoch can create issues if the number of samples grows exponentially.
Therefore, we plot in Figure~\ref{fig:idm_size} the size of $I^s$ for each environment and epoch for the first $450$ epochs.
It is important to note that \abrev{} usually reaches expert performance before its first $100$ epochs.
We observe that for most environments, \abrev{} has a lower slope for appending $I^{pos}$ into its dataset.
This behaviour is excellent since it means that \abrev{} is less likely to create data pool sizes that would transform it to be inefficient.
Furthermore, when we consider that in \citeauthor{TorabiEtAl2018}'s work~\cite{TorabiEtAl2018}, 
$5 \times 10^5$ transitions are needed to learn the inverse dynamic model ($\approx 50$ epochs), this behaviour allows for less preparation when learning an agent.
Nevertheless, Figure~\ref{fig:idm_size} also present two other behaviours.

For the InversePendulum environment, \abrev{} gets almost no sample variation when compared to the other methods in the early stages.
However, after approximately $150$ epochs, $\pi_\theta$ yields trajectories more similar to the expert, which deteriorates $\mathcal{D}$ accuracy and increases $I^s$ quite substantially.
In this environment, we use this behaviour as a form of signal to stop since the reward does not improve. Figure~\ref{fig:idm_size} has an inset graph showing the first $150$ epochs for the InvertedPendulum environment.
In it, we observe during its first epochs, \abrev{} appends samples in a lower rhythm.

For both HalfCheetah and Ant environments, we observe a linear pattern from the samples added into $I^s$.
This behaviour is not desired, resulting in a training procedure that takes around $3$ and $1.5$ times longer to finish than all the other environments for HalfCheetah and Ant.
To mitigate this problem, \abrev{} could implement a forgetting mechanism to get rid of some samples in each epoch either by random selection or using the chronological order of insertion.
However, it should not keep its initial sample pool size, considering it has a smaller dataset and changing it could make $\mathcal{M}$ susceptible to covariate shift.
We hypothesise that adding samples up until an upper limit would be a better approach, eliminating samples from $I^s$ in each epoch as needed to keep the pool size within the limit.

\section{Related Work} \label{sec:related}

The simplest form of imitation learning from observation is Behavioral Cloning (BC)~\citep{Pomerleau1988}, which treats imitation learning as a supervised problem.
It uses samples $(s_t, a, s_{t+1})$ from an expert consisting of a state, action and subsequent state to learn how to approximate the agent's trajectory to the expert's.
However, such an approach becomes costly for more complex scenarios, requiring more samples and information about the action effects on the environment. 
For example, solving Atari requires approximately $100$ times more samples than CartPole.
Generative Adversarial Imitation Learning (GAIL)~\citep{ho2016generative} solves this issue by matching the state-action frequencies from the agent to those seen in the demonstrations, creating a policy with action distributions that are closer to the expert. 
GAIL uses adversarial training to discriminate state-actions either from the agent or the expert while minimising the difference between both.

Recent self-supervised approaches~\citep{TorabiEtAl2018,gavenski2020imitating} that learn from observations use the expert's transitions~$(s^\teacher_t, s^\teacher_{t+1})$ and leverage random transitions $(s_t, a, s_{t+1})$ to learn the inverse dynamics of the environment, and afterwards generate pseudo-labels for the expert's trajectories.
Imitating Latent Policies from Observation (ILPO)~\citep{EdwardsEtAl2019} differs from such work by trying to estimate the probability of a latent action given a state.
Within a limited number of environment steps, it remaps latent actions to corresponding ones.
More recently, Off-Policy Learning from Observations (OPOLO)~\citep{zhu2020off} uses a dual-form of the expectation function and an adversarial structure to achieve off-policy LfO.
Model-Based Imitation Learning from Observation Alone (MobILE)~\citep{kidambi2021mobile} uses the same adversarial techniques, which rely on an objective discriminator coupled with exploration to diverge from its actions when far from the expert.

\section{Conclusions and Future Work}

In this paper, we proposed \method{} (\abrev{}), a new LfO method combining an exploration mechanism and path signatures. 
\abrev{} 
\begin{enumerate*}[label=(\roman*)]
    \item does not require prior domain knowledge or information about the expert's actions; 
    \item has sample efficiency superior or equal to the state-of-the-art LfO alternatives; and 
    \item approximates (sometimes surpassing) expert performance.
\end{enumerate*}
\abrev{} achieves these results due to two key contributions.
Firstly, the use of a discriminator paired with path signatures, allows \abrev{} to acquire more diverse state transition samples while increasing sample quality.
Secondly, the exploration mechanism, which uses the model's error rate to sample from a normal distribution, allows for a more dynamic exploration of the environment.
As a result, the exploration ratio decreases as the model learns to approximate from the ground-truth labels. More importantly, these two innovations are completely model-agnostic, allowing them to be used in other IL methods without requiring major changes.
We would argue that the innovations we proposed pave the way for IL models that generalise better and require less expert training data.

Our next step is to investigate different exploration mechanisms to better fit the policy needs of specific environments. 
We would also like to experiment with different forms of adversarial learning to embed \abrev{}'s current discriminator into the policy loss function. 
Considering the path signatures are differentiable, it would be possible to backpropagate the gradients from the discriminator into the policy.
This change would allow us to see if a direct signal from the enhanced loss function could improve the action prediction of the inverse dynamic model.



\begin{ack}
This work was supported by UK Research and Innovation [grant number EP/S023356/1], in the UKRI Centre for Doctoral Training in Safe and Trusted Artificial Intelligence (\url{www.safeandtrustedai.org}) and made possible via King's Computational Research, Engineering and Technology Environment (CREATE)~\cite{create}.
\end{ack}

\bibliography{references}

\newpage
\appendix
\section{Environments and samples} \label{sup:sec:samples}

\begin{figure*}[!th]
    \centering
    \includegraphics[width=\textwidth]{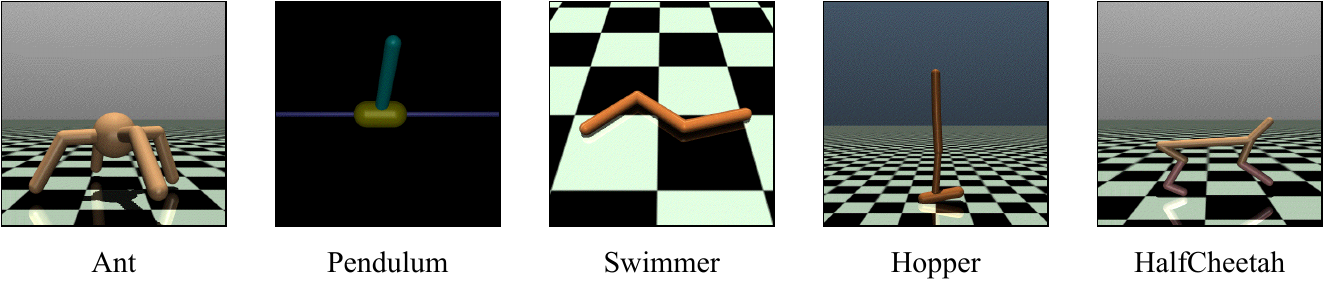}
    \caption{A single frame for each environment used in this work.}
    \label{fig:envs}
\end{figure*}

In this work, we experiment with five different environments.
We now briefly describe each environment and how the expert samples were gathered.
We used Stable Baselines $3$~\cite{stablebaselines3} coupled with RL Zoo$3$~\cite{rlzoo3} to gather expert samples and its weights loaded from HuggingFaces~\footnote{https://huggingface.co/}.
We believe this will facilitate reproducibility by allowing future work to use the exact same experts.
All expert results are displayed in Table~\ref{tab:results} in Section~\ref{sec:sub:results} of the paper.
We used a random sample pool of $50,000$ states for all environments (partitioned into $35,000$ states for training and the remaining $15,000$ for validation).
It is important to note, that in each environment, a dimension $d$ of these state vectors $\vv{v}$ represents an internal attribute of the robot.
Therefore, although they might share a similar number of dimensions, they may carry different meanings.
Since $I^s$ grows in size in each iteration, unlike~\citeauthor{TorabiEtAl2018}'s work~\cite{TorabiEtAl2018}, \abrev{} does not rely on higher sample pools.
Figure~\ref{fig:envs} shows a frame for each environment.

\begin{table*}[ht]
   \centering
   \caption{Layers for each neural network used in this work, where $d$ is the number of dimensions for each state, $\mid a \mid$ is the number of actions, and $\mid \beta \mid$ is given by Eq.~\ref{eq:signature_size}.}
   \label{tab:params}
   \begin{tabular*}{\textwidth}{@{\extracolsep{\fill}}lc@{\extracolsep{\fill}}lc@{\extracolsep{\fill}}lc@{\extracolsep{\fill}}}
        \toprule 
        \multicolumn{2}{c}{$\mathcal{M}$} & \multicolumn{2}{c}{$\agent$} & \multicolumn{2}{c}{$\mathcal{D}$} \\
        \cmidrule{1-2} \cmidrule{3-4} \cmidrule{5-6}
        \multicolumn{1}{c}{Layer Name} & Input $\times$ Output & \multicolumn{1}{c}{Layer Name} & Input $\times$ Output & Layer Name & Input $\times$ Output \\
        Input               & $2d \times 512$           & Input             & $d \times 512$    & Input             & $\mid \beta \mid \times 512$      \\
        Activation (Tanh)   & -                         & Activation (Tanh) & -                 & Activation (Tanh) & -                     \\
        Fully Connected 1   & $512 \times 512$          & Fully Connected 1 & $512 \times 512$  & Fully Connected 1 & $512 \times 512$        \\
        Activation (Tanh)   & -                         & Activation (Tanh) & -                 & Activation (Tanh) & -                     \\
        Self-Attention 1    & $512 \times 512$          & Self-Attention 1  & $512 \times 512$  & Dropout           & 0.5\%                 \\
        Fully Connected 2   & $512 \times 512$          & Fully Connected 2 & $512 \times 512$  & Fully Connected 2 & $512 \times 512$        \\
        Activation (Tanh)   & -                         & Activation (Tanh) & -                 & Activation (Tanh) & -                     \\
        Self-Attention 2    & $512 \times 512$          & Self-Attention 2  & $512 \times 512$  & Dropout           & 0.5\%                 \\
        Fully Connected 3   & $512 \times 512$          & Fully Connected 3 & $512 \times 512$  & Output            & $512 \times 2$        \\
        Activation (Tanh)   & -                         & Activation (Tanh) & -                 & \\
        Fully Connected 4   & $512 \times 512$          & Fully Connected 4 & $512 \times 512$  & \\
        Output              & $512 \times \mid a \mid$          & Output            & $512 \times \mid a \mid$  & \\
        \bottomrule
   \end{tabular*}
\end{table*}

\subsection{A note about the expert samples}
During our experiments, we observed that not all experts are created equally.
Although most experts trained or loaded from HuggingFace share similar results, the behaviour of each expert varies drastically.
One could argue that humans also deviate for each trajectory, but using episodes with a more human-like trajectory (less hectic)
yielded better results for all IL approaches.
By presenting less hectic and more constant movements, we think each policy receives trajectories that vary more and generalise better.
All samples used in this work are available in \url{https://github.com/NathanGavenski/CILO}.

\subsection{Ant-v2}
Ant-v$2$ consists of a robot ant made out of a torso with four legs attached to it, with each leg having two joints~\citep{schulman2015high}. 
The goal of this environment is to coordinate the four legs to move the ant to the right of the screen by applying force on the eight joints.
Ant-v$2$ requires eight actions per step, each limited to continuous values between $-1$ and $1$.
Its observation space consists of $27$ attributes for the $x$, $y$ and $z$ axis of the $3$D robot.
We use Stable Baselines $3$'s TD$3$ weights.
The expert sample contains $10$ trajectories, each with $1,000$ states consisting of $111$ attributes.\footnote{With their 111 different attributes - MuJoCo implementation has $27$ positions with values and the rest with $0$).}
Ant-v$2$ shares distribution behaviour with InvertedPendulum-v$2$, and Hopper-v$2$, having action spaced in a bell-curve. 

\subsection{InvertedPendulum-v2}
This environment is based on the CartPole environment from~\citeauthor{barto1983neuronlike}~\cite{barto1983neuronlike}.
It involves a cart that can move linearly, with a pole attached to it.
The agent can push the cart left or right to balance the pole by applying forces on the cart.
The goal of the environment is to prevent the pole from reaching a particular angle on either side.
The continuous action space varies between $-3$ and $3$, the only one within the five environments outside of the $-1$ to $1$ limit.
Its observation space consists of $4$ different attributes.
We use Stable Baselines~$3$'s PPO weights.
The expert sample size is $10$ trajectories, which consist of $10,000$ states (with their $4$ attributes) and actions (with a single action value per step).
The invertedPendulum-v$2$ environment is the only one that has an expert with the environment's maximum reward.
Therefore achieving $\mathcal{P}$ higher than $1$ is impossible.

\subsection{Swimmer-v2}
This environment was proposed by \cite{coulom2002reinforcement}. It consists of a robot with $s$ segments ($s \geqslant 3$) and $j=s-1$ joints. Following \cite{zhu2020off}, in our experiments we use the default setting $s = 3$ and $j = 2$. 
The agent applies force to the robot's joints, and each action can range from $[-1, 1] \in \mathbb{R}$.
A state is encoded by an $8$-dimensional vector representing the angle, velocity and angular velocity of all segments.
Swimmer distributions present the same distribution of HalfCheetah-v$2$ (centred around the lower and upper limits).
We used Stable Baselines $3$'s TD$3$ weights.
The expert sample contains $4$ trajectories, with $1,000$ states each plus actions for the $j=2$ joints.
The goal of the agent in this environment is to move as fast as possible towards the right by applying torque on the joints and using the fluid's friction.

\subsection{Hopper-v2}
Hopper-v$2$ is based on the work done by \cite{erez2011infinite}.
Its robot is a one-legged two-dimensional body with four main parts connected by three joints: a torso at the top, a thigh in the middle, a leg at the bottom, and a single foot facing the right.
The environment's goal is to make the robot hop and move forward (continuing on the right trajectory).
A state consists of $11$ attributes representing the $z$-position, angle, velocity and angular velocity of the robot's three joints.
We used Stable Baselines $3$'s TD$3$ weights and $10$ expert episodes, each with $1,000$ states and actions for the three joints.
Each action is limited between $[-1, 1] \in \mathbb{R}$.

\subsection{HalfCheetah-v2}
HalfCheetah-v$2$'s environment was proposed in \cite{wawrzynski2009cat}.
It has a $2$-dimensional cheetah-like robot with two ``paws''.
The robot contains $9$ segments and $8$ joints.
Its actions are a vector of $6$ dimensions, consisting of the torque applied to the joints to make the cheetah run forward (``thigh'', ``shin'', and ``paw'' for the front and back parts of the body). 
All states consist of the robot's position and angles, velocities and angular velocities for its joints and segments. 
HalfCheetah-v$2$'s goal is to run forward (i.e., to the right of the screen) as fast as possible. 
A positive reward is allocated based on the distance traversed, and a negative reward is awarded when moving to the left of the screen.
We used Stable Baselines $3$'s TD$3$ weights.
The expert sample size is $10$ trajectories, each consisting of $1,000$ states and actions.
Each action is limited between the interval of $[-1, 1] \in \mathbb{R}$.

\section{Network Topology}
We followed the same network topologies employed in the original works.
Each model ($\mathcal{M}$ and $\pi_\theta$) are MLP with 4 fully connected layers, each with 512 neurons, with the exception of the last layer whose size is the same as the number of environment actions, Table~\ref{tab:params} displays the topologies alongside the input and output sizes of each layer.
Following the implementation in \cite{gavenski2020imitating}, we used a self-attention module after the first and second layers.
We experimented with normalisation layers during development, which did not increase the agents' results but helped with weight updates.
Although we understand that having more complex architectures could increase our method's performance, for consistency we used the same original architecture to show that \abrev{} achieves expert results and does not rely on the architecture.
The implementation of our method can be found within \url{https://github.com/NathanGavenski/CILO}.

\section{Training and Learning Rate}
For training, we used a Nvidia A$100$ $40$GB GPU and PyTorch.
Although we used this GPU, such hardware is not strictly required since \abrev{} uses $\approx 2$GB to train with a $1024$ mini-batch size.
The learning rates for $\mathcal{M}$ and $\agent$ are shown in Table~\ref{tab:learning_rate}.
We note that \abrev{} is robust to different learning rates for $\pi_\theta$.
However, $\mathcal{M}$ is more sensitive since $I^s$ changes at almost every iteration, assuming there is at least one agent's trajectory that $\mathcal{D}$ classifies as expert.
Having a high learning rate can make $\mathcal{M}$'s weights update too harshly and result in \abrev{} never learning how to label the $\Tau^{\teacher}$ properly.

\begin{table}[ht]
    \centering
    \caption{Different learning rates for $\mathcal{M}$ and $\pi_\theta$ for all environments.}
    \label{tab:learning_rate}
    \begin{tabular*}{\columnwidth}{@{\extracolsep{\fill}}lccc}
        \toprule Environment & $\mathcal{M}$ & $\pi_\theta$ & Signature $k$  \\
        \midrule Ant & $1\times 10^3$ & $1\times 10^3$ & 2\\ 
        \midrule InvertedPendulum & $1\times 10^3$ & $1\times 10^3$ & 4\\ 
        \midrule Swimmer & $3\times 10^3$ & $7\times 10^4$ & 4 \\
        \midrule Hopper & $5\times 10^3$ & $1\times 10^3$ & 4 \\ 
        \midrule HalfCheetah & $1\times 10^3$ &  $7\times 10^4$ & 4 \\  
        \bottomrule
    \end{tabular*}
\end{table}

\section{Path Signatures}

In this work we rely on several path signature definitions to discriminate over agent and expert trajectories.
In Section~\ref{sec:sub:goal} of our paper, we briefly defined a trajectory $\tau$, in which each state is a vector $\vv{v}$ in $\mathbb{R}^d$, and how to compute the path signature $\beta(\tau)_{1, n}^{i_1, \cdots. i_k}$, where $n$ is the length of the trajectory, and $i_1, \cdots, i_k \in \{1, \cdots, d \}$ ($k>1$) are indices to elements in $\vv{v}$.
Here, we provide some additional information on the process of computing a path signature and the intuition behind it.

\subsection{Computing the Path Signature} \label{sec:comp_sig}

Given a trajectory $\tau$ and a function $f$ that interpolates $\tau$ into a continuous map $f: \mathbb{R} \rightarrow \mathbb{R}$, the integral of the the trajectory against $f$ can be defined as:
\begin{equation}
    \int_1^n  f(\tau_t)d\tau_t = \int_1^n f(\tau_t)\dot{\tau}_tdt,
\end{equation}
where $\dot{\tau}_t = \frac{d\tau_t}{dt}$ for any time $t \in [1, n]$.
Note that $f(\tau_t)$ is a real-valued path defined on $[1, n]$, which can be considered the integral of a trajectory $\tau$.
Moreover, if we consider that $f(\tau_t) = 1$ for all $t \in [1, n]$, then the path integral of $f$ against any trajectory $\tau: [1, n] \rightarrow \mathbb{R}$ is simply the increment of $\tau$:
\begin{equation}
    \int_1^n d\tau_t = \int_1^n \dot{\tau}dt = \tau_n - \tau_1.    
\end{equation}

Therefore, by assuming that $\beta$ is a function of real-valued paths, we can define the signature for any single index $i_k \in \{1, \cdots, d \}$ as:
\begin{equation} \label{eq:single_index}
    \beta(\tau)^{i_k}_{1, n} = \int_{1 < s \leqslant n} d\tau^{i_k}_s = \tau^{i_k}_n - \tau^{i_k}_1,
\end{equation}
which is the increment of the $i_k$-th dimension of the path.
Now, if we move to any pair of indexes $i_k, j_k \in \{1, \cdots, d \}$, we have to consider the double-iterated integral:
\begin{equation} \label{eq:double_index}
    \beta(\tau)^{i_k, j_k}_{1, n} = \int_{1 < s \leqslant n} \beta(\tau)^{i_k}_{1, s}d\tau^{j_k}_s = \int_{1 < r \leqslant s \leqslant n} d\tau^{i_k}_t d\tau^{j_k}_s,
\end{equation}
where $\beta(\tau)^{i_k}_{1, s}$ is given by Eq.~\ref{eq:single_index}.
Considering that $\beta(\tau)^{i_k, j_k}_{1, n}$ continues to be a real-values path, then we can define recursively the signature function for any number of indexes $k \geqslant 1$ in the collection of indexes $i_1, \cdots, i_k \in \left\{ 1, \cdots, d\right\}$ as:
\begin{equation} \label{eq:signature}
    \beta(\tau)^{i_1,\cdots,i_k}_{1, n} =
    \int_{1 < s \leqslant n} \beta(\tau)^{i_1, \cdots, i_{k-1}}_{1, s} d\tau^{i_k}_s,
\end{equation}
which in our paper is Eq.~\ref{eq:recursive_signature}. 
It is important to note that $k$ is the depth up to which the signature is generated (not its length). 
At each level $i \leqslant k$, ``words'' of length $i$ are generated from the alphabet $D$ according to Eq.~\ref{eq:recursive_signature} (main work) to produce the terms of the signature.
Figure~\ref{fig:dictionary} shows all possible terms for a trajectory $\tau: [1, n] \rightarrow \mathbb{R}^d$ for different depths.
For example, a signature with depth $2$ will have all the terms in the levels $i=0,1,2$. In an alphabet with $d$ letters, we can construct one word of length $0$, $d$ words of length $1$, and $d^2$ words of length $2$, giving $1+d+d^2$ words in total (i.e., the number of terms in the signature). 
In general, the length of a signature with alphabet size $d$ and depth $k$ is:
\begin{equation}\label{eq:signature_size}
\sum_{i=0}^{k}d^i=\frac{d^{k+1}-1}{d-1}.
\end{equation}
We observe that signatures can be computed for any depth $k$, and are not restricted to $k \leqslant d$.

We give two examples to illustrate how the terms in a signature are computed (we omit the level $0$ whose single value $1$ is fixed).
Figure~\ref{fig:signatures_single} shows how to generate a signature of depth $1$ (with the terms in the first and second columns of Figure~\ref{fig:dictionary}).
Considering that the length of a signature grows exponentially with the depth $k$ desired ($\frac{d^{k+1}-1}{d-1}$), Figure~\ref{eq:double_index} only shows how to calculate the terms of a signature of depth $2$ for a $2$-dimensional dictionary, with the first index fixed in $2$.

\begin{figure}[ht]
   \centering 
   \subfloat[$\beta(\tau)^{i}_{1,n} \mid \forall_i \in \{1, \cdots, d \}$.\label{fig:signatures_single}]{
        \begin{tikzpicture}[]

    \matrix (m) [
        matrix of nodes,
        nodes in empty cells,
        row sep=0pt,
        column sep=0pt,
        nodes={minimum width=20pt, minimum height=20pt}
    ]{
        $s^1_1$ & $s^2_1$ & $\cdots$ & $s^d_1$ \\
        $s^1_2$ & $s^2_2$ & $\cdots$ & $s^d_2$ \\
        $\vdots$ & $\vdots$ & $\vdots$ & $\vdots$ \\
        $s^1_n$ & $s^2_n$ & $\cdots$ & $s^d_n$ \\
    };

    \draw[xshift=-4pt, yshift=0pt] 
        ([xshift=5pt]m-1-4.north) --
        (m-1-4.north east) --
        (m-4-4.south east) --
        ([xshift=5pt]m-4-4.south) {};
    \node [above=1pt] at (m-1-1.north)  {\scriptsize $1$};
    \node [above=1pt] at (m-1-4.north)  {\scriptsize $d$};
    \node [above=1pt] at (m-1-2.north east)  {\scriptsize $\cdots$};

    \draw[xshift=0pt, yshift=-4pt] 
        ([xshift=-5pt]m-1-1.north) --
        (m-1-1.north west) -- 
        (m-4-1.south west) --
        ([xshift=-5pt]m-4-1.south) {};
    \node [left=1pt] at (m-1-1.west)  {\scriptsize $1$};
    \node [left=1pt] at (m-4-1.west)  {\scriptsize $n$};
    \node [left=1pt] at (m-2-1.south west) {\scriptsize $\vdots$};

    \draw [draw=cb_blue]
        ([xshift=-2.5pt, yshift=-3pt]m-1-1.north) -- 
        ([xshift=3.5pt, yshift=-3pt]m-1-1.north west) -- 
        ([xshift=3.5pt, yshift=3pt]m-4-1.south west) --
        ([xshift=-2.5pt, yshift=3pt]m-4-1.south){};

    \draw [draw=cb_blue]
        ([xshift=2.6pt, yshift=-3pt]m-1-1.north) -- 
        ([xshift=-3.5pt, yshift=-3pt]m-1-1.north east) -- 
        ([xshift=-3.5pt, yshift=3pt]m-4-1.south east) --
        ([xshift=2.5pt, yshift=3pt]m-4-1.south){};
    \node [below=1pt, text=cb_blue] at (m-4-1.south) {\fontsize{4}{4}\selectfont $\beta(\tau)^1_{1, n}$};

    \draw [draw=cb_orange]
        ([xshift=-2.5pt, yshift=-3pt]m-1-2.north) -- 
        ([xshift=3.5pt, yshift=-3pt]m-1-2.north west) -- 
        ([xshift=3.5pt, yshift=3pt]m-4-2.south west) --
        ([xshift=-2.5pt, yshift=3pt]m-4-2.south){};

    \draw [draw=cb_orange]
        ([xshift=2.6pt, yshift=-3pt]m-1-2.north) -- 
        ([xshift=-3.5pt, yshift=-3pt]m-1-2.north east) -- 
        ([xshift=-3.5pt, yshift=3pt]m-4-2.south east) --
        ([xshift=2.5pt, yshift=3pt]m-4-2.south){};
    \node [below=1pt, text=cb_orange] at (m-4-2.south) {\fontsize{4}{4}\selectfont $\beta(\tau)^2_{1, n}$};
    
    \draw [draw=cb_green]
        ([xshift=-2.5pt, yshift=-3pt]m-1-4.north) -- 
        ([xshift=3.5pt, yshift=-3pt]m-1-4.north west) -- 
        ([xshift=3.5pt, yshift=3pt]m-4-4.south west) --
        ([xshift=-2.5pt, yshift=3pt]m-4-4.south){};

    \draw [draw=cb_green]
        ([xshift=2.6pt, yshift=-3pt]m-1-4.north) -- 
        ([xshift=-3.5pt, yshift=-3pt]m-1-4.north east) -- 
        ([xshift=-3.5pt, yshift=3pt]m-4-4.south east) --
        ([xshift=2.5pt, yshift=3pt]m-4-4.south){};
    \node [below=1pt, text=cb_green] at (m-4-4.south) {\fontsize{4}{4}\selectfont $\beta(\tau)^d_{1, n}$};
\end{tikzpicture}
    }
   \hfill
   \subfloat[$\beta(\tau)^{2, j}_{1,n} \mid \forall_j \in \{1, \cdots, d \}$.\label{fig:signatures_double}]{
        \begin{tikzpicture}[]

    \matrix (m) [
        matrix of nodes,
        nodes in empty cells,
        row sep=0pt,
        column sep=0pt,
        nodes={minimum width=20pt, minimum height=20pt}
    ]{
        $s^1_1$ & $s^2_1$ & $\cdots$ & $s^d_1$ \\
        $s^1_2$ & $s^2_2$ & $\cdots$ & $s^d_2$ \\
        $\vdots$ & $\vdots$ & $\vdots$ & $\vdots$ \\
        $s^1_n$ & $s^2_n$ & $\cdots$ & $s^d_n$ \\
    };

    \draw[xshift=-4pt, yshift=0pt] 
        ([xshift=5pt]m-1-4.north) --
        (m-1-4.north east) --
        (m-4-4.south east) --
        ([xshift=5pt]m-4-4.south) {};
    \node [above=1pt] at (m-1-1.north)  {\scriptsize $1$};
    \node [above=1pt] at (m-1-4.north)  {\scriptsize $d$};
    \node [above=1pt] at (m-1-2.north east)  {\scriptsize $\cdots$};

    \draw[xshift=0pt, yshift=-4pt] 
        ([xshift=-5pt]m-1-1.north) --
        (m-1-1.north west) -- 
        (m-4-1.south west) --
        ([xshift=-5pt]m-4-1.south) {};
    \node [left=1pt] at (m-1-1.west)  {\scriptsize $1$};
    \node [left=1pt] at (m-4-1.west)  {\scriptsize $n$};
    \node [left=1pt] at (m-2-1.south west) {\scriptsize $\vdots$};

    \draw [draw=cb_blue]
        ([xshift=-2.5pt, yshift=-3pt]m-1-1.north) -- 
        ([xshift= 3.5pt, yshift=-3pt]m-1-1.north west) -- 
        ([xshift= 3.5pt, yshift= 3pt]m-4-1.south west) --
        ([xshift=-2.5pt, yshift= 3pt]m-4-1.south){};

    \draw [draw=cb_blue]
        ([xshift= 2.5pt, yshift=-3pt]m-1-1.north) -- 
        ([xshift=-3.5pt, yshift=-3pt]m-1-1.north east) -- 
        ([xshift=-3.5pt, yshift= 3pt]m-4-1.south east) --
        ([xshift= 2.5pt, yshift= 3pt]m-4-1.south){};
        
    \draw [draw=cb_blue]
        ([xshift=-2.5pt, yshift=-2pt]m-1-2.north) -- 
        ([xshift=2.5pt, yshift=-2pt]m-1-2.north west) -- 
        ([xshift=2.5pt, yshift=2pt]m-4-2.south west) --
        ([xshift=-2.5pt, yshift=2pt]m-4-2.south){};

    \draw [draw=cb_blue]
        ([xshift=2.6pt, yshift=-2pt]m-1-2.north) -- 
        ([xshift=-2.5pt, yshift=-2pt]m-1-2.north east) -- 
        ([xshift=-2.5pt, yshift=2pt]m-4-2.south east) --
        ([xshift=2.5pt, yshift=2pt]m-4-2.south){};
    \node [below=1pt, text=cb_blue] at (m-4-1.south) {\fontsize{4}{4}\selectfont $\beta(\tau)^{2, 1}_{1, n}$};
    
    \draw [draw=cb_orange]
        ([xshift=-2.5pt, yshift=-3pt]m-1-2.north) -- 
        ([xshift=3.5pt, yshift=-3pt]m-1-2.north west) -- 
        ([xshift=3.5pt, yshift=3pt]m-4-2.south west) --
        ([xshift=-2.5pt, yshift=3pt]m-4-2.south){};

    \draw [draw=cb_orange]
        ([xshift=2.6pt, yshift=-3pt]m-1-2.north) -- 
        ([xshift=-3.5pt, yshift=-3pt]m-1-2.north east) -- 
        ([xshift=-3.5pt, yshift=3pt]m-4-2.south east) --
        ([xshift=2.5pt, yshift=3pt]m-4-2.south){};
    \node [below=1pt, text=cb_orange] at (m-4-2.south) {\fontsize{4}{4}\selectfont $\beta(\tau)^{2, 2}_{1, n}$};
        
    \draw [draw=cb_green]
        ([xshift=-2.5pt, yshift=-1pt]m-1-2.north) -- 
        ([xshift= 1.5pt, yshift=-1pt]m-1-2.north west) -- 
        ([xshift= 1.5pt, yshift= 1pt]m-4-2.south west) --
        ([xshift=-2.5pt, yshift= 1pt]m-4-2.south){};
        
    \draw [draw=cb_green]
        ([xshift=-2.5pt, yshift=-3pt]m-1-4.north) -- 
        ([xshift=3.5pt, yshift=-3pt]m-1-4.north west) -- 
        ([xshift=3.5pt, yshift=3pt]m-4-4.south west) --
        ([xshift=-2.5pt, yshift=3pt]m-4-4.south){};

    \draw [draw=cb_green]
        ([xshift= 2.6pt, yshift=-1pt]m-1-2.north) -- 
        ([xshift=-1.5pt, yshift=-1pt]m-1-2.north east) -- 
        ([xshift=-1.5pt, yshift= 1pt]m-4-2.south east) --
        ([xshift= 2.5pt, yshift= 1pt]m-4-2.south){};

    \draw [draw=cb_green]
        ([xshift=2.6pt, yshift=-3pt]m-1-4.north) -- 
        ([xshift=-3.5pt, yshift=-3pt]m-1-4.north east) -- 
        ([xshift=-3.5pt, yshift=3pt]m-4-4.south east) --
        ([xshift=2.5pt, yshift=3pt]m-4-4.south){};
    \node [below=1pt, text=cb_green] at (m-4-4.south) {\fontsize{4}{4}\selectfont $\beta(\tau)^{2, d}_{1, n}$};
\end{tikzpicture}
    } \\
   \subfloat[Collection of signatures for $\tau$, where $k \in [0, \infty)$. \label{fig:dictionary}]{
        \begin{tikzpicture}[]

    
    \matrix (m) [
        matrix of nodes,
        nodes in empty cells,
        row sep=0pt,
        column sep=3pt,
        nodes={minimum width=20pt, minimum height=20pt}
    ]{
        1 & $\beta(\tau)^1_{1, n}$ & $\beta(\tau)^{1, 1}_{1, n}$ & $\beta(\tau)^{1, 1, 1}_{1, n}$ & $\cdots$ & $\beta(\tau)^{1, 1, \cdots, 1}_{1, n}$ \\
          & $\beta(\tau)^2_{1, n}$ & $\beta(\tau)^{1, 2}_{1, n}$ & $\beta(\tau)^{1, 1, 2}_{1, n}$ & $\cdots$ & $\beta(\tau)^{1, 1, \cdots, 2}_{1, n}$ \\
          & $\vdots$               & $\vdots$                    & $\vdots$                       &          & $\vdots$ \\
          & $\beta(\tau)^d_{1, n}$ & $\beta(\tau)^{1, d}_{1, n}$ & $\beta(\tau)^{1, 1, d}_{1, n}$ & $\cdots$ & $\beta(\tau)^{1, 1 \cdots, d}_{1, n}$ \\
          &                        & $\beta(\tau)^{2, 1}_{1, n}$ & $\beta(\tau)^{1, 2, 1}_{1, n}$ & $\cdots$ & $\beta(\tau)^{i_1, i_2, \cdots, i_k}_{1, n}$ \\
          &                        & $\vdots$                    & $\vdots$                       &          & $\vdots$ \\
          &                        & $\beta(\tau)^{d, 1}_{1, n}$ & $\beta(\tau)^{d, d, 1}_{1, n}$ & $\cdots$ & $\beta(\tau)^{d, d, \cdots, 1}_{1, n}$ \\
          &                        & $\beta(\tau)^{d, 2}_{1, n}$ & $\beta(\tau)^{d, d, 2}_{1, n}$ & $\cdots$ & $\beta(\tau)^{d, d, \cdots, 2}_{1, n}$ \\
          &                        & $\vdots$                    & $\vdots$                       &          & $\vdots$ \\
          &                        & $\beta(\tau)^{d, d}_{1, n}$ & $\beta(\tau)^{d, d, d}_{1, n}$ & $\cdots$ & $\beta(\tau)^{d, d, \cdots, d}_{1, n}$ \\
    };
    
    \draw [->, >=latex]
        ([xshift=-8pt, yshift=2pt]m-1-1.north west) -- 
        ([yshift=2pt]m-1-6.north east) {};
    \node [above=2pt] at (m-1-1.north)  {\scriptsize $0$};
    \node [above=2pt] at (m-1-2.north)  {\scriptsize $1$};
    \node [above=2pt] at (m-1-3.north)  {\scriptsize $2$};
    \node [above=2pt] at (m-1-4.north)  {\scriptsize $3$};
    \node [above=2pt] at (m-1-5.north)  {\scriptsize $\cdots$};
    \node [above=2pt] at (m-1-6.north)  {\scriptsize $k$};

    \draw [->, >=latex]
        ([yshift=10pt]m-1-1.north west) --
        (m-10-1.south west) node[midway, left=2pt] {\rotatebox{90}{$d^i\text{ items}$}};
    \node [xshift=-6pt, above=2pt] at (m-1-1.north west) {\scriptsize $i$};
    \node [xshift=10pt, above=12pt] at (m-1-4.north west) {\scriptsize depth};
\end{tikzpicture}
    }
   \caption{Illustration of path signature. \label{fig:signatures}}
\end{figure}

\subsection{A Numerical Example}

\begin{figure}[ht]
    \centering
    \includegraphics[width=.35\textwidth]{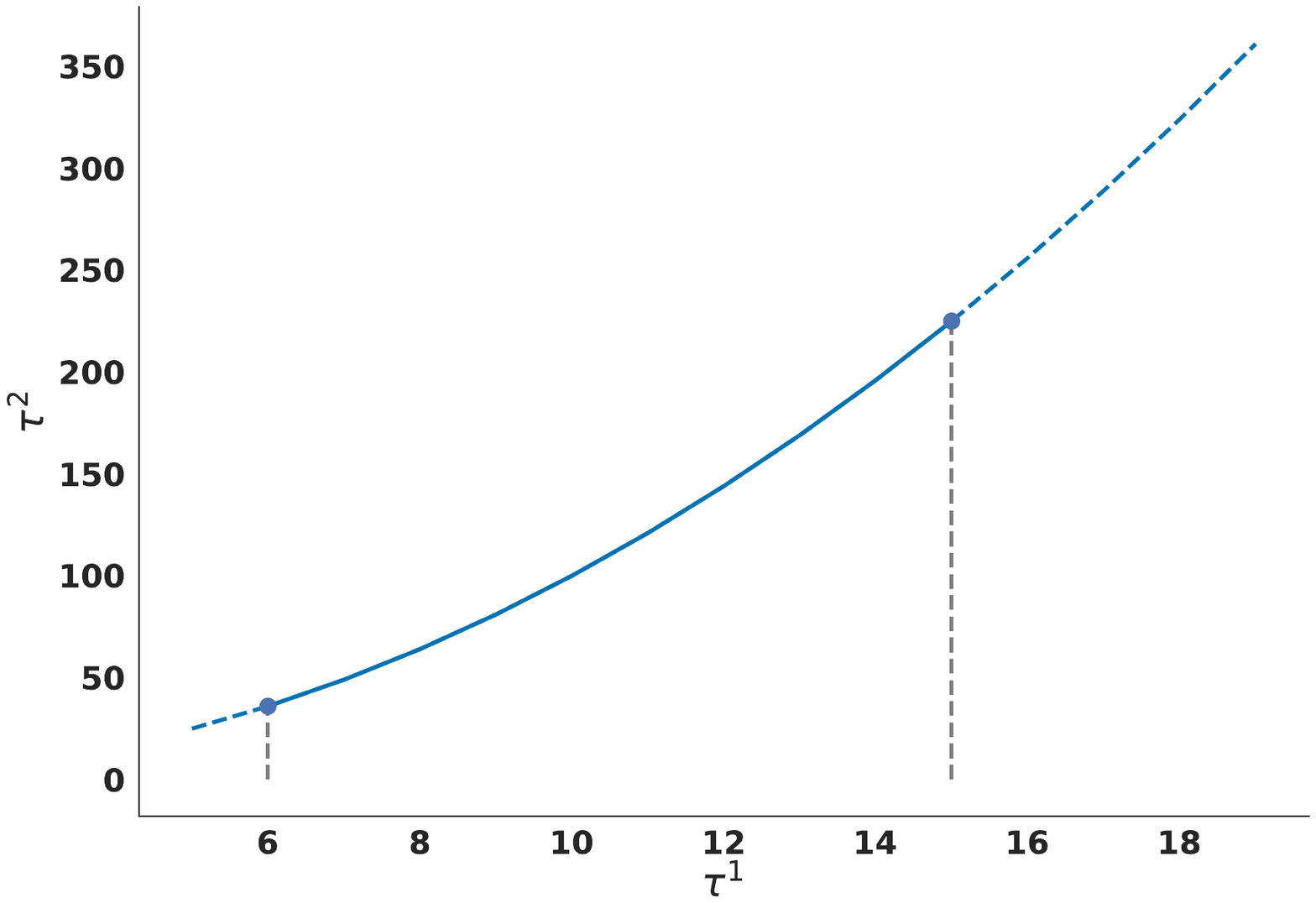}
    \caption{Trajectory $\tau: [1, 10] \rightarrow \mathbb{R}^2$.}
    \label{fig:example}
\end{figure}

Let us consider a trajectory $\tau$ with two two-dimensional states $\{\tau_t^1, \tau_t^2 \}$, and the set of multi-indexes $W = \left\{\right. \left( i_1, \cdots, i_k\right) \mid k \geqslant 1, i_1, \cdots, i_k \in \left\{1, 2\right\} \left.\right\}$, which is the set of all finite sequences of $1$'s and $2$'s.
Given the trajectory  $\tau: [1, 10] \rightarrow \mathbb{R}^2$ illustrated in Figure~\ref{fig:example}, where the path function for $\tau$ is computed according to the function:
\begin{equation}
    \begin{aligned}
        \tau_t &= \{ \tau_t^1, \tau_t^2 \} = \{ 5 + t, (5 + t)^2 \mid t \in [1, 10] \} \\
    \end{aligned}
\end{equation}

For the depth $k$ desired, the computation of the signature would be computed as shown in Figure~\ref{fig:step-by-step}. For example, given that states in $\tau$ are two-dimensional ($d=2$), the path signature for $\tau$ with depth $k=2$ will have the $\frac{d^{k+1}-1}{d-1}=\frac{2^{3}-1}{1}=7$ terms in the vector $\beta(\tau)_{1, 10} = \left[\right. 1,  9, 189, 40.5, 970.5, 730.5, 17860.5 \left.\right]$.

\begin{figure}
    \centering
    \begin{tikzpicture}[scale=1.5, node distance=2mm, every node/.style={align=center, anchor=west}]
    \node[align=center] (k3) {
        $\begin{aligned}
            \beta(\tau)^{1, 1, 1}_{1, 10} &= \int\!\!\!\!\int\!\!\!\!\int_{1 < t_1 \leqslant t_2 \leqslant t_3 \leqslant 10} d\tau^1_{t_1}d\tau^1_{t_2}d\tau^1_{t_3} = \int_1^{t_1} \left[ \int_1^{t_2} \left[ \int_1^{t_3} dt_1 \right] dt_2 \right] dt_3 = 121.5 \\
            &\vdots
        \end{aligned}$
    };

    \node[align=center, above=of k3] (k2) {
        $\begin{aligned}
            \beta(\tau)^{1, 1}_{1, 10} &= \int\!\!\!\!\int_{1 < t_1 \leqslant t_2 \leqslant 10} d\tau^1_{t_1}, d\tau^1_{t_2} = \int_1^{10} \left[ \int_1^{t_2} dt_1 \right] dt_2 = 40.5 \\
            \beta(\tau)^{1, 2}_{1, 10} &= \int\!\!\!\!\int_{1 < t_1 \leqslant t_2 \leqslant 10} d\tau^1_{t_1}, d\tau^2_{t_2} = \int_1^{10} \left[ \int_1^{t_2} dt_1 \right] 2(5 + t_2)dt_2 = 970.5 \\
            \beta(\tau)^{2, 1}_{1, 10} &= \int\!\!\!\!\int_{1 < t_1 \leqslant t_2 \leqslant 10} d\tau^2_{t_1}, d\tau^1_{t_2} = \int_1^{10} \left[ \int_1^{t_2} 2(5+t)dt_1 \right] dt_2 = 730.5 \\
            \beta(\tau)^{2, 2}_{1, 10} &= \int\!\!\!\!\int_{1 < t_1 \leqslant t_2 \leqslant 10} d\tau^2_{t_1}, d\tau^2_{t_2} = \int_1^{10} \left[ \int_1^{t_2} 2(5+t)dt_1 \right] 2(5+t)dt_1 = 17,860.5
        \end{aligned}$
    };
    
    \node[above=of k2.north west, anchor=south west] (k1) {
        $\begin{aligned}
            \beta(\tau)^1_{1, 10} &= \int_{1 < t \leqslant 10} dt = \tau^1_{10} - \tau^1_1 = 9\\
            \beta(\tau)^1_{1, 10} &= \int_{1 < t \leqslant 10} dt = \tau^2_{10} - \tau^2_1 = 189
        \end{aligned}$
    };
    
    \node[above=of k1.north west, anchor=south west] (k0) {$\beta(\tau)^{\o}_{1, n} = 1$};

    \draw[-, >=latex] ([yshift=-2pt]$(current bounding box.west |- k0.south west)$) -- ([yshift=-2pt]$(current bounding box.east |- k0.south east)$);
    \draw[-, >=latex] ([yshift=-2pt]$(current bounding box.west |- k1.south west)$) -- ([yshift=-2pt]$(current bounding box.east |- k1.south east)$);
    \draw[-, >=latex] ([yshift=-2pt]$(current bounding box.west |- k2.south west)$) -- ([yshift=-2pt]$(current bounding box.east |- k2.south east)$);

    \node[anchor=east] at ([xshift=-2pt]k0.west) {\rotatebox{90}{\scriptsize $k = 0$}};
    \node[anchor=east] at ([xshift=-2pt]k1.west) {\rotatebox{90}{\scriptsize $k = 1$ -- Eq. 3}};
    \node[anchor=east] at ([xshift=-2pt]k2.west) {\rotatebox{90}{\scriptsize $k = 2$ -- Eq. 4}};
    \node[anchor=east] at ([xshift=-2pt]k3.west) {\rotatebox{90}{\scriptsize $k \geqslant 3$ -- Eq. 5}};

\end{tikzpicture}
    \caption{Step-by-step computation of a path signature\label{fig:step-by-step}}
\end{figure}

\subsection{Signature Properties}                                                             
We now describe properties of path signatures that are most relevant to our work.
The description is not comprehensive. We recommend the work from~\citeauthor{yang2022developing}~\cite{yang2022developing} and~\citeauthor{chevyrev2016primer}~\cite{chevyrev2016primer} for  a more in-depth approach to path signatures.

\hfill \break
\noindent
\textbf{Uniqueness:}
This property relates to the fact that no two trajectories $\tau$ and $\tau'$ of bounded variation have the same signature unless the trajectories are tree-equivalent.
In light of the invariance under reparametrisations~\cite{lyons2014rough}, we note that path signatures have no tree-like sections to monotone dimensions, such as acceleration.

\hfill \break
\noindent
\textbf{Generic nonlinearity of the signature:}
 The second property refers to the product of two terms $\beta(\tau)^{i_1, \cdots, i_k}$ and $\beta(\tau)^{j_1, \cdots, j_k}$, which can also be expressed as:
 \begin{equation}
    \begin{aligned}
         \beta(\tau)^1_{1, n} \cdot \beta(\tau)^{2}_{1, n} &= \beta(\tau)^{1, 2}_{1, n} + \beta(\tau)^{2, 1}_{1, n} \text{ or}, \\
         \beta(\tau)^{1, 2}_{1, n} \cdot \beta(\tau)^{1}_{1, n} &= \beta(\tau)^{1, 1, 2}_{1, n} + \beta(\tau)^{1, 2, 1}_{1, n}.
    \end{aligned}
 \end{equation}
Thus, the nonlinearity of the signature in terms of low-level terms can be expressed by the linear combination of higher-level terms, which adds more nonlinear previous knowledge to the feature vector.
This behaviour is better exemplified in the second level of signatures where for any $\beta(\tau)^{i_k, i_k}_{1, n}$, the result will be $\nicefrac{\left( \tau_n^{i_k} - \tau_1^{i_k} \right)^2}{2}$.

\hfill \break
\noindent
\textbf{Fixed dimension under length variations:}
The last property refers to the path signature's length invariance under different trajectory lengths.
In Section~\ref{sec:comp_sig}, we showed that the signature length is a function of the signature depth ($k$) and the number of dimensions in a state ($d$).
Therefore, path signatures become practical feature vectors for trajectories in machine learning tasks, requiring different inputs to share the same size without recurrent neural networks.

\subsection{Motivation for Signatures}
Given the nature of deep learning methods operating on vectorial data, which requires the input data to be of a predetermined fixed length, many techniques, such as word embeddings (where a word is represented by a vector), are used to circumvent this length requirement.
Moreover, imitation learning tasks, by definition, have to effectively represent expert demonstrations to capture relevant information for learning a desired behaviour.
Path signatures provide a solution to represent sequential or trajectory-based expert demonstration in a principled and efficient manner.

In imitation learning, expert demonstration often takes the form of trajectories or sequences of states over time.
Path signatures offer a way to encode these trajectories into high-dimensional feature representations that capture the expert behaviour in a geometric and analytic way.
Furthermore, the uniqueness property ensures that essential information about the expert demonstrations is preserved in the path signature representation, enabling accurate discrimination over different trajectories' signatures.
Lastly, path signatures provide a single hyperparameter (the number of desired collections $k$).
By adjusting $k$, we can control the trade-off between representational quality and computational complexity, allowing for efficient learning and generalisation.
However, we observe that increasing $k$ leads to an exponential increase in the length of the signature, which imposes a limit to agents with limited computation resources.

\subsection{Signature Time Complexity}

\begin{figure}
    \centering
    \includegraphics[width=\columnwidth]{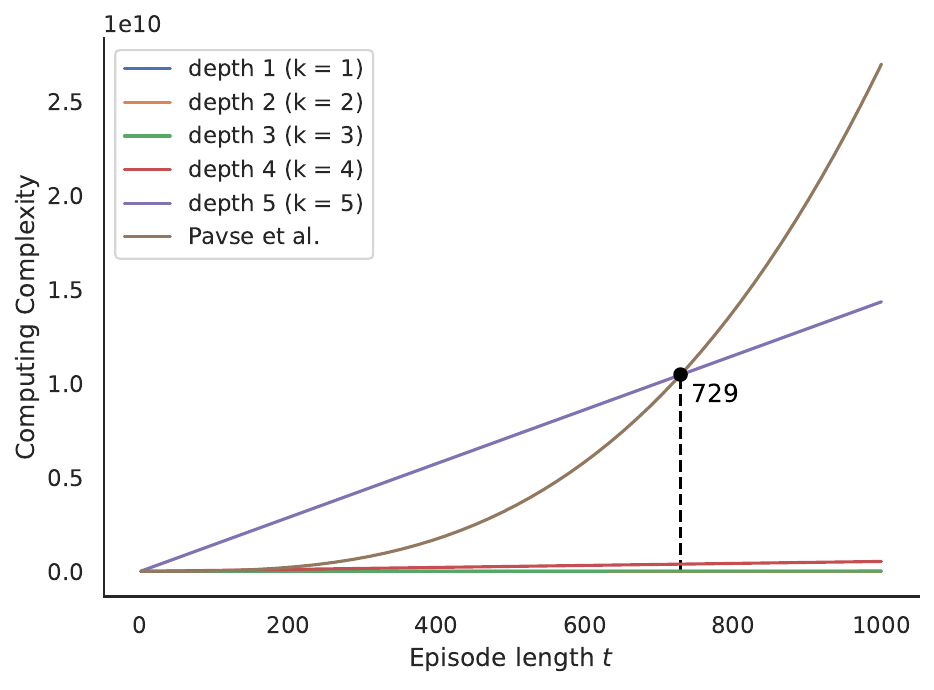}
    \caption{Upper-bound time complexity for signature computation and average.}
    \label{fig:complexity}
\end{figure}

We now briefly discuss the upper-bound complexity for computing path signatures and compare it to \citeauthor{pavse2020ridm}'s work~\cite{pavse2020ridm}, which computes the averages of the trajectory states.
Given Eq.~\ref{eq:signature} and Fig.~\ref{fig:signatures}, it should be easy to see that path signatures can be computed in time $\mathcal{O}(t \cdot d^k)$, where $t$ is the number of samples in a trajectory, $d$ is the number of dimensions, and $k$ is the depth of the signature.
In contrast, \citeauthor{pavse2020ridm}'s work~\cite{pavse2020ridm} uses the average over the current and previous states. This does not work well when different trajectories average to the same value, but it is not an issue for signatures due to their uniqueness.
Using averages is not an issue in \citeauthor{pavse2020ridm}'s work (or in IRL in general), which computes an artificial reward signal at each timestep. 
However, it also quickly becomes costly because the method computes the average of all sub-trajectories $t$ times at each epoch and trajectories are traversed multiple times ($\mathcal{O}(d \cdot t^3)$).
The cost of computing signatures increases linearly with respect to the episode length, whereas the cost of computing \citeauthor{pavse2020ridm}'s averages increases exponentially.
Moreover, path signatures increase exponentially according to the number of dimensions $d$, which is constant for all environments.
The main parameter affecting the cost of computing signatures in \abrev{} is the depth $k$. Fig.~\ref{fig:complexity} compares the costs for Ant-v$2$ --- the environment with the largest state representation ($d=111$). The figure shows that the cost of computing signatures is lower than that of computing averages for signatures with depth up to $5$. In Ant-v$2$ (the environment with the highest number of dimensions) with signature depths up to $5$ and episode length at least $729$, the cost of computing signatures is lower than computing \citeauthor{pavse2020ridm}'s averages. Moreover, recall we only needed to use depth $2$ to obtain superior performance.

\end{document}